\documentclass[manuscript,screen]{acmart}
\AtBeginDocument{%
  \providecommand\BibTeX{{%
    \normalfont B\kern-0.5em{\scshape i\kern-0.25em b}\kern-0.8em\TeX}}}

\setcopyright{acmcopyright}
\copyrightyear{2018}
\acmYear{2018}
\acmDOI{XXXXXXX.XXXXXXX}
\acmJournal{TKDD}

\usepackage{multirow}
\usepackage[labelformat=simple]{subcaption}

\usepackage{makecell}
\usepackage{siunitx}
\usepackage{enumitem}
\usepackage{colortbl}
\usepackage{mathtools}
\usepackage{adjustbox}
\usepackage{makecell}
\usepackage[ruled]{algorithm2e} 
\definecolor{mygray}{gray}{.9}
\newcolumntype{M}[1]{>{\centering\arraybackslash}m{#1}}
\newcommand*{\model}{{CSST}\@\xspace}

\begin{document}
%
\title{Spatio-Temporal Contrastive Self-Supervised Learning for POI-level Crowd Flow Inference}
%
%
%
%

\author{Songyu Ke}
\authornote{This research was done when the first author was an intern at JD Intelligent Cities Research \& JD iCity under the supervision of Junbo Zhang.}
\email{songyu-ke@outlook.com}
\orcid{0000-0001-7184-8074}
\affiliation{%
  \institution{Shanghai Jiao Tong University}
  \city{Shanghai}
  \country{China}
  \postcode{200240}
}
\affiliation{
    \institution{JD Intelligent Cities Research}
    \city{Beijing}
    \country{China}
}
\affiliation{
    \institution{JD iCity, JD Technology}
    \city{Beijing}
    \country{China}
}
\author{Ting Li}
\email{liting6259@gmail.com}
\author{Li Song}
\email{song200626@gmail.com}
\author{Yanping Sun}
\email{sunyanping7@jd.com}
\author{Qintian Sun}
\email{sqt51515@gmail.com}
\author{Junbo Zhang}
\authornote{Junbo Zhang is the corresponding author.}
\email{msjunbozhang@outlook.com}
\orcid{0000-0001-5947-1374}
\author{Yu Zheng}
\email{msyuzheng@outlook.com}
\affiliation{
    \institution{JD Intelligent Cities Research}
    \city{Beijing}
    \country{China}
}
\affiliation{
    \institution{JD iCity, JD Technology}
    \city{Beijing}
    \country{China}
}
\renewcommand{\shortauthors}{Songyu~Ke, et al.}
\begin{abstract}
Accurate acquisition of crowd flow at Points of Interest (POIs) is pivotal for effective traffic management, public service, and urban planning. Despite this importance, due to the limitations of urban sensing techniques, the data quality from most sources is inadequate for monitoring crowd flow at each POI. This renders the inference of accurate crowd flow from low-quality data a critical and challenging task. The complexity is heightened by three key factors: 1) \emph{The scarcity and rarity of labeled data}, 2) \emph{The intricate spatio-temporal dependencies among POIs}, and 3) \emph{The myriad correlations between precise crowd flow and GPS reports}.
    
To address these challenges, we recast the crowd flow inference problem as a self-supervised attributed graph representation learning task and introduce a novel \underline{C}ontrastive \underline{S}elf-learning framework for \underline{S}patio-\underline{T}emporal data (\model). Our approach initiates with the construction of a spatial adjacency graph founded on the POIs and their respective distances. We then employ a contrastive learning technique to exploit large volumes of unlabeled spatio-temporal data. We adopt a swapped prediction approach to anticipate the representation of the target subgraph from similar instances. Following the pre-training phase, the model is fine-tuned with accurate crowd flow data. Our experiments, conducted on two real-world datasets, demonstrate that the \model pre-trained on extensive noisy data consistently outperforms models trained from scratch.
\end{abstract}


\maketitle
\section{Introduction}
Monitoring the crowd flow at POIs (Points of Interest) is essential for a modern city. For urban management, fine-grained POI crowd flow monitoring can help people better detect abnormal events (e.g., traffic accidents) in the city and evacuate people in time to prevent dangers caused by excessive gatherings. For enterprises, it can also support marketing, advertising, and other aspects to increase their exposure among the target population. 

In the past years, with the development of location-acquisition technology, huge amounts of human mobility data have been accumulated, which benefits a wide range of intelligent city applications \cite{zheng2014urban,zheng2019urban}. Some excellent works have recently been proposed to predict the future crowd flow at POIs based on historical values and other attributes. \cite{li2016mflow} proposes a method to estimate the crowd flow of an area based on the cellular network data. \cite{STResNet,lin2019deepstn} divides the whole city into grids, computes the crowd flow by counting traffic trajectories, and employs convolutional networks to predict the future crowd flow. Moreover, \cite{junkai2020} extends the crowd flow prediction to irregular regions by introducing graph networks. 

\begin{figure}[tb]
    \centering
    \includegraphics[width=\linewidth]{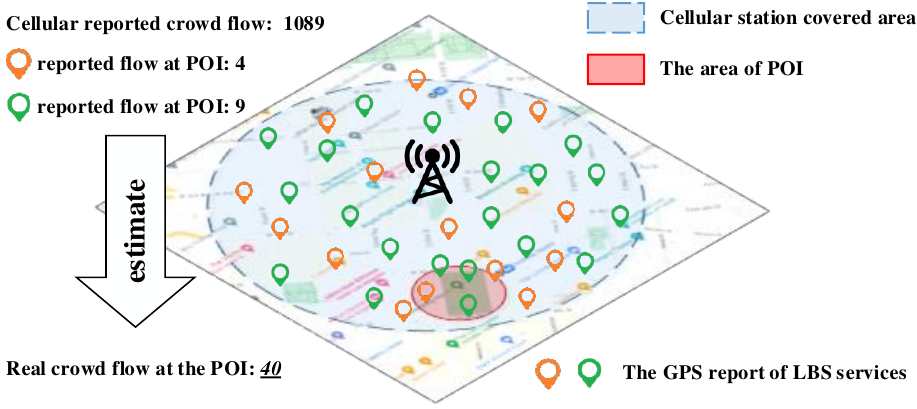}
    \caption{Issues when estimating the crowd flow by cellular signaling data or users' GPS reports.}
    \label{fig:issue-low-quality-data}
\end{figure}

However, collecting accurate crowd flow data at each POI is almost impossible to achieve.
Due to the limitations of sensing devices, the human mobility data from cellular signaling data and location-based service (LBS) is usually of low quality. More specifically, as shown in Figure \ref{fig:issue-low-quality-data}, the cellular signaling data is spatially coarse that cannot discriminate which POI the user accessed actually. Meanwhile, since user preferences, the GPS reports from location-based service providers are usually biased and incomplete. 

\begin{figure}[t]
    \centering
    \subcaptionbox
        {Proportions\label{City}}
        [0.6\linewidth]
        {\includegraphics[width=\linewidth]{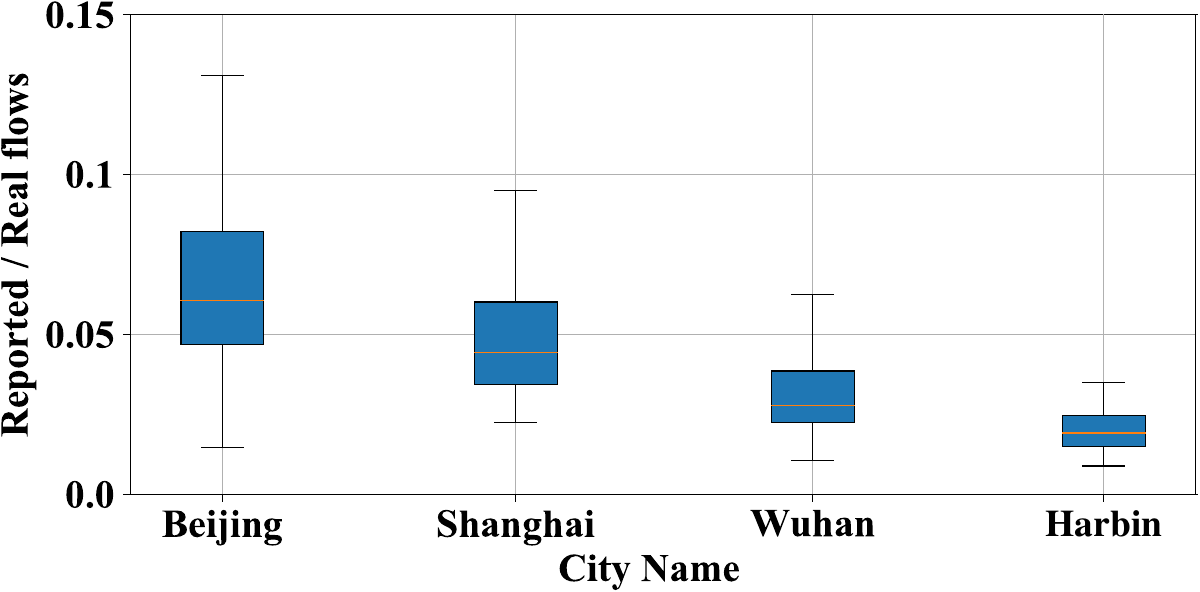}}
    \subcaptionbox
        {Impact of area\label{Area}}
        [0.48\linewidth]
        {\includegraphics[width=\linewidth]{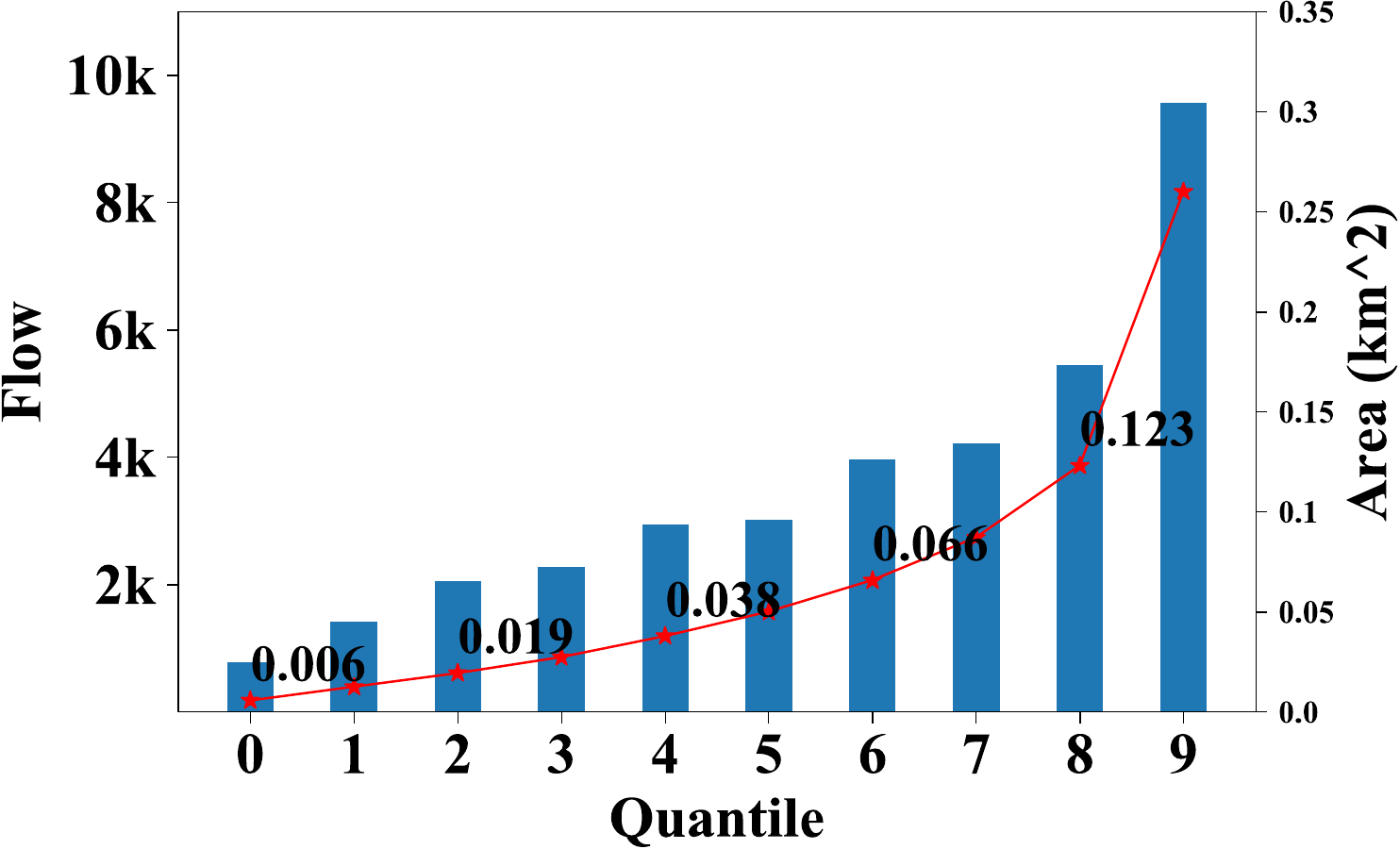}}
    \hfill
    \subcaptionbox
        {Impact of profile\label{Profile}}
        [0.48\linewidth]
        {\includegraphics[width=\linewidth]{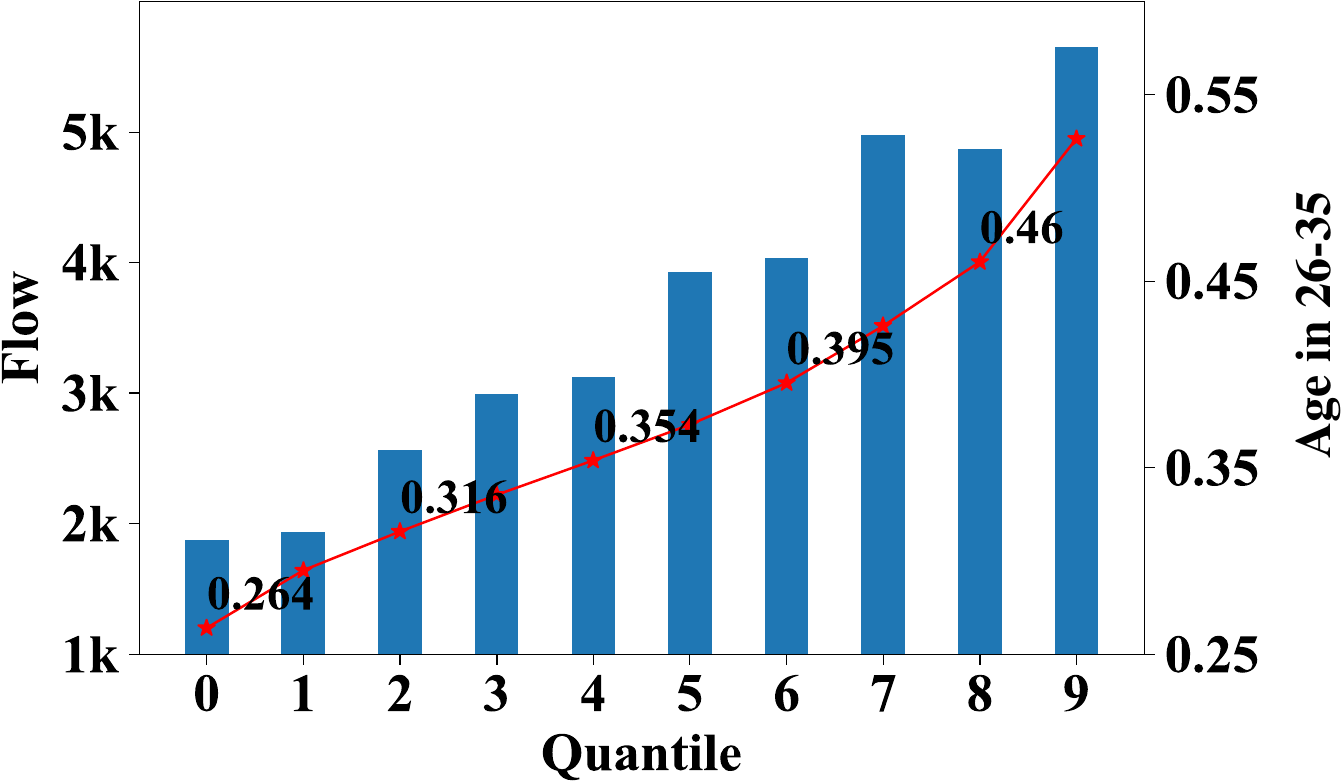}}
    \caption{Illustration of proportion in different cities (a) and the influence of area and profile (b) and (c). \label{intro:data_analysis}}
\end{figure}

Therefore, there are three challenges in monitoring crowd flows at POIs: 
\begin{enumerate}[leftmargin=*]
    \item \emph{The precise POI crowd flow data is rare, sparse, and expensive.} As mentioned above, cellular signaling data and GPS reports cannot describe the crowd flow at POIs accurately. Therefore, the accurate crowd flow (i.e., labels) can only be computed by gathering multiple data sources, including cellular signaling data and GPS reports from different LBSs. Moreover, such a process involves the acquisition, cleaning, and aggregation of multi-party data, which requires a lot of manual effort and domain knowledge, and is difficult to obtain on a large scale.
    \item \emph{Spatio-temporal dependencies between POIs are non-trivial.} The crowd flow of one POI may be affected by its neighbors since people may move from one area to another. The influence may be dynamic due to different social activities. People go to office buildings on weekdays and hang out in parks at weekends. In addition, POIs with similar attributes may have similar flow distribution. 
    \item \emph{The relationship between GPS reports and the actual flow is complicated.} Figure \ref{City} presents the proportion and its distribution of the GPS reports to the actual flow in four key cities, and the median ratios in four cities are all less than $0.1$, indicating that the crowd flow inferred by location-based service is severely missing. Moreover, even in the same city, the proportions of different POIs vary greatly. The real POI crowd is affected by complex factors (\textit{e.g., \,} the area, transportation condition, and crowd portraits). Figure \ref{Area} and \ref{Profile}, where the horizontal axis is the feature quantile, show the real flow change with the area and young adults' proportion. We can observe that the crowd flow fluctuated and increased with more young adults and larger areas.
\end{enumerate}

Therefore, we need a method to recover the accurate POI crowd flow by exploiting large-scale low-quality data sources. More specifically, we are trying to learn the correlation between low-quality GPS reports and POI features with accurate POI crowd flow from a few supervised labels and then infer fine-grained crowd flow on a large scale. The key to solving the crowd flow inference problem is \emph{how to learn a transferable representation}.

\begin{figure*}[tb]
    \centering
    \includegraphics[width=\linewidth]{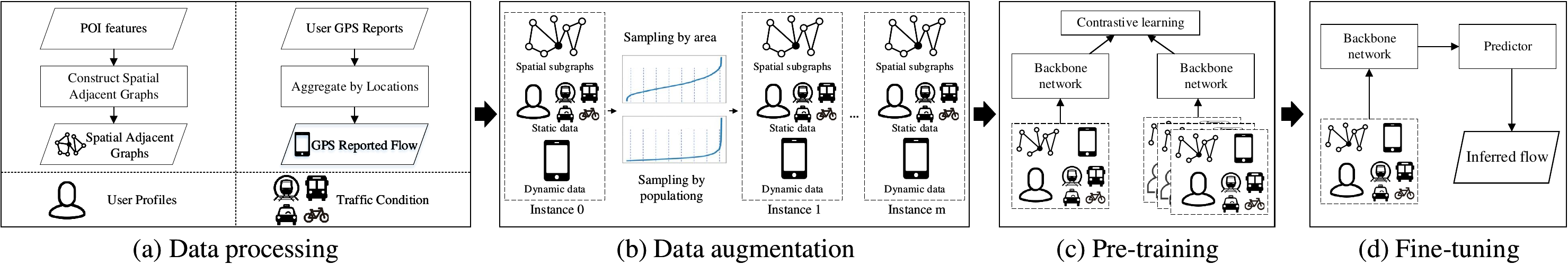}
    \caption{The framework of \model. (a) Data processing: constructing spatial adjacent graphs from the features of nearby POIs and computing the GPS reported flow by aggregating user GPS reports; (b) Data augmentation: discovering similar instances and building positive sample pairs for spatio-temporal representation learning; (c) Pre-training: learning the spatio-temporal representation via contrastive learning; (d) Fine-tuning: use rare labeled data to fine-tune the network.}\label{intro:framework}
\end{figure*}

To tackle the challenges above, we propose a framework termed \model (\underline{C}ontrastive \underline{S}elf-learning framework for \underline{S}patio-\underline{T}emporal data), as shown in Figure \ref{intro:framework}, to model the correlation of low-quality GPS reports and external factors with the real crowd flow. The first step of the framework is data processing. We collect the static information including nearby POI features, traffic conditions, user profiles and the dynamic series, i.e., the GPS reported flow aggregated from the raw GPS reports (as Figure \ref{intro:framework}(a) shown). The rest part of the proposed framework consists of three main steps. We first divide the POIs into several categories based on the key attributes and then sample the positive pairs in the same category. Then, we exploit a spatio-temporal representation network to encode the heterogeneous features and a contrastive learning framework to optimize the attributed graph embedding on huge amounts of unlabeled data. Finally, we use a small amount of labeled data to fine-tune the representation on the POI flow inference task. Unlike the existing contrastive learning framework, in this paper, we mainly focus on designing the effective spatio-temporal representation network and the subgraph sampling strategy based on region attributes to generate the data augmentations of the target instance.

To the best of our knowledge, this is the first approach that generalizes contrastive learning to the fine-grained ST flow inference problem. Our contributions can be summarized into the following three aspects:
\begin{itemize}[leftmargin=*]
    \item We propose a novel contrastive self-supervised learning method for spatio-temporal data named \model for POIs' flow inference which introduces contrastive learning to solve the problem of the lack of accurate flow data (i.e., labels). Furthermore, \model is task-agnostic and tailored for spatio-temporal data.
    \item We propose an effective spatio-temporal data augmentation strategy to generate similar instance pairs for spatio-temporal data and exploit the swapped contrastive coding to optimize the model. Moreover, we applied the proposed self-supervised learning framework to a spatio-temporal representation network to solve the POI flow inference problem, which consists of fully connected layers to learn the representation of each category of features, respectively, and a message-passing graph network to learn the spatial adjacency dependency.
    \item We perform extensive experiments with three deep backbone networks on two kinds of POIs. The experimental results demonstrate that \model can significantly improve the performance on a small number of labeled data.
\end{itemize}

\section{Preliminaries}
This section briefly introduces the definitions and the spatial-temporal prediction problem statement. For brevity, the frequently used notations in this paper are presented in Table \ref{tab:notation}.

\begin{table}[tb]
	\centering
	\caption{Notations.}
	\label{tab:notation}
	\begin{tabular}{l|l}
		\Xhline{1pt}
		\textbf{Notations} & \textbf{Description}\\
		\Xhline{1pt}
        $N, k$ & the number of POIs and neighbors\\
        \hline
		$m$ & the number of data augmentations\\
		\hline
		$\mathcal{G}=\{\mathcal{V}, \mathcal{E}\}$ & the attributed adjacent grpah\\
		\hline
		$\mathcal{S}$ & the set of data augmentations\\
		\hline
		$\mathcal{N}(i)$ & the set of neighbors of POI $l_i$\\
		\hline
        $\mathbf{x}^{i}$ & the set of GPS reported users of POI $l_i$\\
        \hline
        $\mathbf{v}_{i}$ & the attributes of POI $l_i$\\
        \hline
        $\mathbf{v}_{i}^c$ & the crow portrait features of POI $l_i$\\
        \hline
        $\mathbf{v}_{i}^a$ & the area, traffic and locations features of POI $l_i$\\
        \hline
        $D_a$ & the dimension of the area, traffic, and locations \\
        \hline
        $D_c$ & the dimension of crow portrait features\\
        \hline
        $D_r$ & the dimension of GPS report features \\
        \hline
        $L_m$ & the number of MLP layers \\
        \hline
		$L_c$ & the number of graph convolution layers\\
		\Xhline{1pt}
	\end{tabular}
\end{table}
\newtheorem{definition_problem}{Definition}
\begin{definition_problem}
		\textbf{Attributed Adjacent Graph:}
		An attributed graph is defined as $\mathcal{G}=\{\mathcal{V}, \mathcal{E}\}$, where $\mathcal{V}$ denotes the set of $N$ POIs, and $\mathcal{E}$ denotes the set of edges between POIs. In this paper, we consider the k-nearest neighbors for each POI and $d_{i,j} \in \mathbb{R}$ is the geospatial distance between POI $l_i$ and $l_j$. Furthermore, $\mathbf{v}_{i} \in \mathbb{R}^{D_a + D_c}$ denotes the $D_a+D_c$-dimension attributes of POI $l_i$, where $D_a$ is the dimension of inherent attributes $v_i^a$ such as area, transportation condition, and location, and the $D_c$ is the dimension of crowd portrait features $v_i^c$ consisting of the proportion of each age group and each gender.
\end{definition_problem}

\begin{definition_problem}
		\textbf{GPS Reports:}
		We collected the mobile app reports associated with longitude-latitude information and count the number of people in the polygon of each POI. We denote the GPS reported users in POI $l_i$ at time $t$ as $\mathbf{x}_{t}^{i}$.
\end{definition_problem}

\textbf{Problem Statement: } Given POI $l_i$, let the observed GPS reported users in $i$ at time $t$ be $\mathbf{x}_{t}^{i}$, spatial adjacency graph is constructed as $\mathcal{A}_{i}$ where each node $l_j$ in $\mathcal{A}_{i}$ is associated with traffic condition $\mathbf{v}^{a}_j$ and crowd portrait $\mathbf{v}^{c}_j$, and each edge represents the distance $d_{ij}$ to neighbor $l_j$, our target is to infer the accurate crowd flow for POI $l_i$ at the current timestamp $\mathbf{y}_{t}^{i}$.

\section{Methodology}
This section presents the details of the contrastive learning framework for spatio-temporal data, \model.

\subsection{Contrastive Self-Supervised Learning}
Our spatio-temporal contrastive framework \model contains two steps:
1) node attribute-based data augmentation;
2) the node representation computing and the optimization of the backbone network.
In this section, we will introduce these two steps in detail.

\subsubsection{Data Augmentation}\label{sec:data_aug}
For a contrastive self-supervised learning framework, data augmentation plays an important role in learning effective representation. 
In computer vision\cite{mview2019contrastive, wu2018unsupervised}, people usually transform the source image by cropping, resizing, color jittering, or flipping randomly and combine these augmented instances with the source image as similar image pairs. As for graph representation learning, GCC\cite{qiu2020gcc} utilizes random walks to sample paths on the graph, generating multiple instances starting from a particular node.

However, for the POI flow inference task, the adjacency graph $\mathcal{A}$ is constructed with geospatial distance, indicating that pairwise similarity decreases as the distance increases. Furthermore, the POI attributes, such as the area and online reported users, also contribute to the crowd flow. Therefore, the random walk on a graph may introduce noise due to the weak relationship between distances and GPS reports.

\begin{figure}[t]
    \centering
    \includegraphics[width=\linewidth]{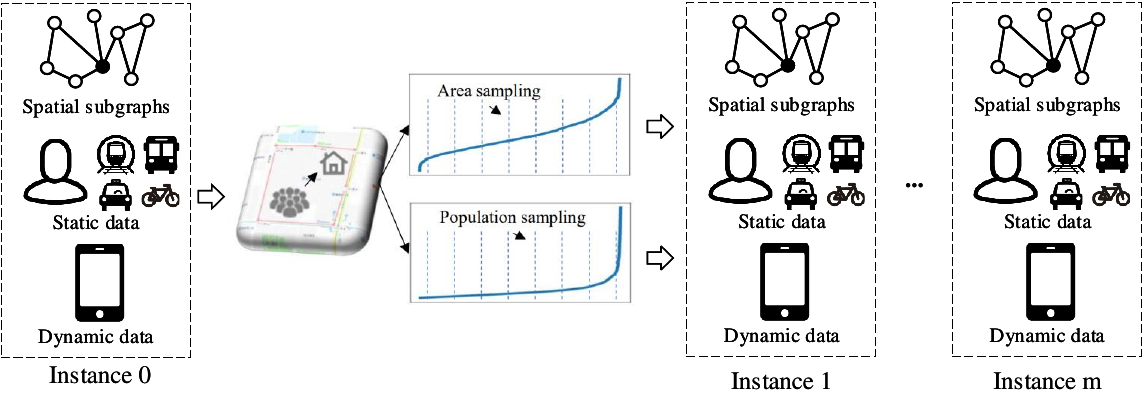}
    \caption{The spatio-temporal data augmentation method of \model.}\label{method:data_aug}
\end{figure}

In \model, we propose to generate similar subgraph instances by using two POI inherent attributes, i.e., the area and the GPS reports, as shown in Figure \ref{method:data_aug}. First, we compute the quantile of each attribute, then separate it into several sets of intervals. After that, for target POI $l_i$ in the interval $I$, we sample $m$ instances in the same area interval and the same GPS report interval as our targets.
In our opinion, POI pairs have similar attributes \textit{e.g.,} areas or GPS reports, which means similar crowd flow. That can be formulated as:
\begin{equation}\label{data_sample}
    \mathcal{S}=\{\mathbf{l}_{s_1}, ..., \mathbf{l}_{s_{m}}\} \subseteq \mathcal{S}^{a} \cap \mathcal{S}^{r}
\end{equation}
where $m$ is the number of similar instances, $\mathcal{S}^{a}$ is the set of instances from the same area interval as the target, and $\mathcal{S}^{r}$ is the set of instances in the same GPS report interval
Different from the instance in computer vision, which is a single image or video, a spatio-temporal instance contains the k-hop adjacent graph of $l_{s_i}$, the static attributes, and the GPS reports.
 
\begin{figure*}[!t]
    \centering
    \subcaptionbox
        {\small Backbone network\label{method_backbone}}
        {\includegraphics[width=0.43\linewidth]{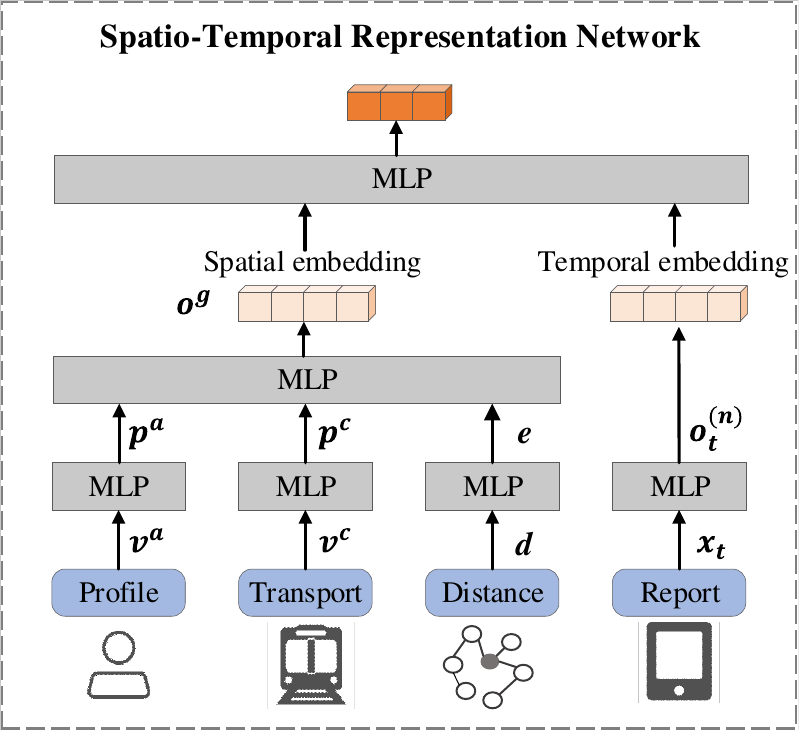}}
    \subcaptionbox
        {\small Attributed graph contrastive framework of \model\label{method_swap}} 
        {\includegraphics[width=0.54\linewidth]{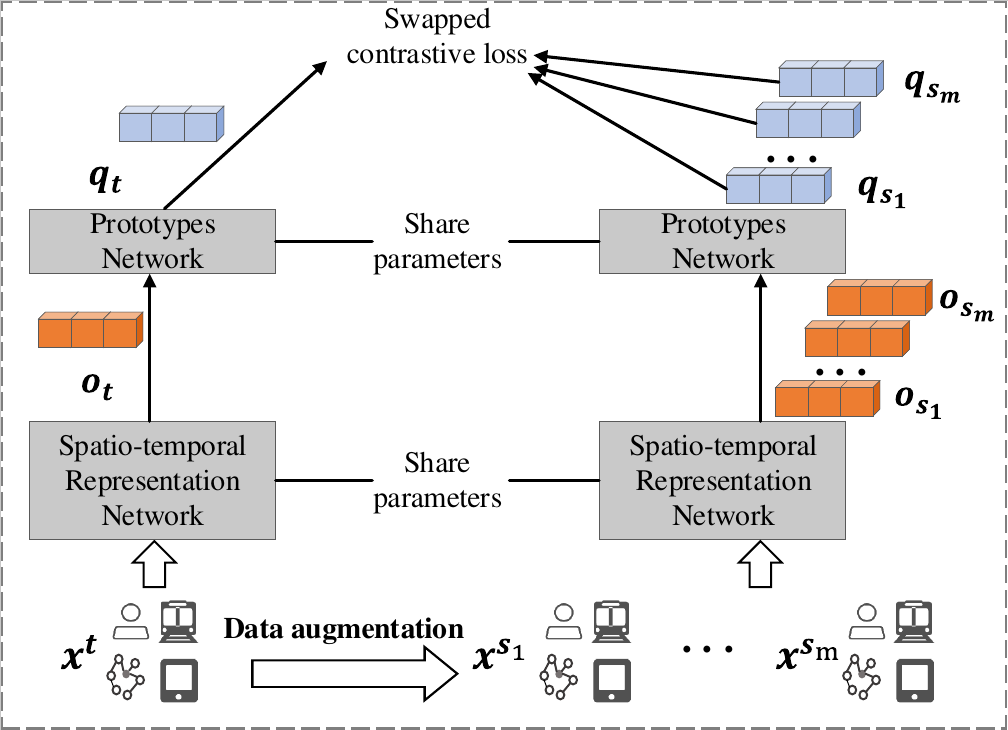}}
    \caption{The backbone network and contrastive framework of CSST(a) the backbone network which encode the spatio-temporal heterogeneous features; (b) the attributed graph contrastive framework where prototypes network share the same parameters.}
    \label{fig:model}
\end{figure*}

\subsubsection{Pre-Training}\label{sec:pre-train}
Recently, self-supervised learning frameworks, consisting of pre-training and fine-tuning, have achieved great success in computer vision\cite{he2019moco} and natural language processing\cite{devlin2018bert, glue2018, GPT01, LM01}. This paper explores learning high-quality representations using large-scale low-quality datasets composed of area, crowd portraits, traffic conditions, and noisy GPS reports. Precisely, we follow the same contrastive strategy as \cite{caron2021swaped}, using a swapped prediction mechanism to predict the value of one view from the representation of another view. The swapped framework can be trained with small batches, which means better memory efficiency than existing methods\cite{he2019moco, chen2020bigcsl, qiu2020gcc}.
The prototype network in Figure \ref{method_swap} is a fully connected network, which shares parameters between source and augmented instances. Moreover, the swapped contrastive loss is defined in Eq.(\ref{main_swap_loss}):
\begin{equation}\label{main_swap_loss}
    \mathcal{L}(\mathbf{o}_{t}, \mathbf{o}_{s}) = \mathcal{L}(\mathbf{o}_{t}, \mathbf{q}_{s}) + \mathcal{L}(\mathbf{q}_{t}, \mathbf{o}_{s})
\end{equation}
where $\mathbf{o}_{t}$ and $\mathbf{o}_{s}$ are outputs of backbone networks,  $\mathbf{q}_{s}$ and $\mathbf{q}_{t}$ are outputs of prototypes networks.
The $\mathcal{L}$ is the cross entropy loss between the representation and the probability which is computed by taking a softmax of the dot production of $\mathbf{o}_{t}$ and all prototypes in $\mathbf{C}$ :
\begin{equation}\label{swap_loss}
    \begin{gathered}
        \mathcal{L}(\mathbf{o}_{t}, \mathbf{q}_{s})= - \sum_{k=1}^{K} \mathbf{q}_{s}^{(k)} \log \mathbf{p}_{t}^{(k)}  \\
        \text{where} ~ \mathbf{p}_{t}^{k} = \frac{\exp(\frac{1}{\tau} \mathbf{o}_{t}^{\top} \mathbf{c}_{k})}{\sum_{k^{'}} \exp(\frac{1}{\tau} \mathbf{o}_{t}^{\top} \mathbf{c}_{k^{'}})}
    \end{gathered}
\end{equation}
where $\tau$ is the temperature parameter\cite{wu2018unsupervised}, $K$ is the number of prototypes. Moreover, we follow the same solution in SWaV \cite{caron2021swaped}, which restricts the transportation of tensors in the mini-batch to ensure that the model is memory efficient.
\subsection{Spatio-Temporal Representation Networks}\label{method_stgnn}
The crowd flow of POIs is affected by complicated factors. Therefore, we use a multi-source fusion network to encode these heterogeneous features, and Figure \ref{method_backbone} shows its architecture. The encoder network of \model considers four kinds of features. The first is the spatial adjacency graph $\mathcal{A}_{i}$, where the edge $d_{ij}$ is the distance to neighbors. And there are two categories of features for each node: 1) the inherent attributes of POIs $\mathbf{v}^{a}$, which contains area transportation and the location; 2) the POIs' crowd portraits $\mathbf{v}^{c}$, which consisting of the proportions of people of all ages and all genders. The most important feature is the GPS reports $\mathbf{x}_{t}^{i}$ collected by mobile apps, which are dynamic at different timestamps.

We first exploit two multilayer perceptron (MLP) networks to encoder two kinds of POI attributes, respectively, which is formulated as Eq.(\ref{node_encoder}):
\begin{equation}\label{node_encoder}
    \begin{gathered}
        \mathbf{p}_{i}^{0} = \mathbf{p}_{i}^{a} \oplus \mathbf{p}_{i}^{c} \\
        \text{where}~\mathbf{p}_{i}^{a} = f_a(\mathbf{v}_{i}^{a};\theta^{a}), \mathbf{p}_{i}^{c} = f_c(\mathbf{v}_{i}^{c};\theta^{c})
    \end{gathered}
\end{equation}
where $\oplus$ is the concatenate operation, $\mathbf{p}_{i}^{0}$ denotes the embedding of POI $l_i$ which is the $0$-th layer input of graph neural network,
$\mathbf{v}_{i}^{a}$ and $\mathbf{v}_{i}^{c}$ are inputs of embedding network $f_a$ and $f_c$ which represent the inherent spatial attributes of POI $l_i$,
and $\mathbf \theta^{a}$ and $\mathbf \theta^{c}$ are trainable parameters of these two networks.
Then, inspired by the message passing neural network (MPNN)\cite{mpnn2017icml}, which considers the weights of edges to learn the node representation, we define the edge network, which learns the mapping function from weights to the embedding space as Eq.(\ref{edge_encoder}):
\begin{equation}\label{edge_encoder}
\begin{gathered}
   \mathbf{e}_{ij} = f_e(w_{ij};\theta^{e}) \\
   \text{where}~w_{ij} = \exp(-{d_{ij}^{2}}/{\sigma^{2}})
\end{gathered}
\end{equation}
where $w_{ij}$ is the input of the edge network $f_e$, and $\theta^{e}$ is the learnable parameters.
To be specific, we compute the weight of edge $w_{ij}$ based on the adjacent distance $d_{ij}$, which is the geo-spatial distance between POI $l_i$ and $l_j$.
Moreover, $\sigma$ is a hyper-parameter to adjust the exponential attenuation ratio.

To learn the representation of the graph structure, we exploit the MPNN\cite{mpnn2017icml} to consider the edge weights.
\emph{Note that MPNN can be replaced by a fully connected network, which can be applied to \model and achieve good performance when the label data is sufficient.}
MPNN contains three main components: a message-passing network $M$, a vertex-updating network $U$, and a readout network $R$.
Among them, the message-passing network learns the message from neighbor nodes with edge features,
the vertex-updating network updates the hidden state of each node based on the output of the message-passing network,
and the readout network generates a feature vector for the whole $k$-hop adjacent graph. The whole MPNN can be defined as Eq.~(\ref{message_network}):
\begin{equation}\label{message_network}
    \mathbf{o}^{g} = \sum_{l_j \in \mathcal{N}(l_i)} f_{g}(\mathbf{p}_{i}, \mathbf{p}_{j}, \mathbf{e}_{ij};\theta_{g})
\end{equation}
where $\mathbf{p}_i$ denotes the output of fully connected layer, POI $l_j$ is the neighbor of $l_i$, $\mathcal{N}(l_i)$ deontes the neighbor set of POI $l_i$,
$\mathbf{e}_{ij}$ is output of the edge mapping function $f_e$.
Moreover, $\theta^{g}$ is the parameters of the readout network $f_g$.

Finally, we exploit a fusion network to model the relationship between noisy GPS reports $\mathbf{x}_{t}^{i}$ and the static spatial features in graph structure $\mathbf{o}^{g}$. We use a MLP network to encode the noisy GPS reports $\mathbf{x}_{t}^{i}$ shown in Eq.(\ref{report_network}), then adopt a fusion layer (defined in \ref{fusion_network}) to jointly model these two kinds of features:
\begin{equation}\label{report_network}
    \mathbf{o}_{t}^{(n)} = f_n(\mathbf{x}_{t}^{i}; \theta^{n})
\end{equation}
\begin{equation}\label{fusion_network}
    \mathbf{o}_{t} = f_s(\mathbf{o}_{t}^{(n)} \oplus \mathbf{o}^{(g)}; \theta^{s})
\end{equation}
where $\mathbf{o}_{t}^{(n)}$ is the output of noisy GPS reports network $f_n$, which has the learnable parameters $\theta^{n}$.
Also, the concatenate of noisy GPS reporting features $\mathbf{o}_{t}^{n}$ and graph structure embedding $\mathbf{o}^{(g)}$ is the input of fusion network $f_s$, which has trainable parameters $\theta^{s}$.

\subsubsection{\model Fine-Tuning}
After learning the representation with large-scale noisy data, we fine-tune the task-specific features on a small amount of labeled data.
Specifically, we feed the \emph{task-agnostic} features into a regression network to infer the accurate crowd flow:
\begin{equation}\label{output network}
    \mathbf{\hat{y}}_{t} = sigmoid(f_o(\mathbf{o}_{t}; \theta_o))
\end{equation}
where $\mathbf{o}_{t}$ is the output of the backbone network, $\theta_o$ is the trainable parameters of regression network $f_o$. Furthermore, \emph{sigmoid} is a nonlinear activation function. Finally, we use a binary cross-entropy function to compute the loss.
In this paper, we adopt the full-tuning strategy that the backbone network is initialized with pre-trained parameters and then trained end-to-end together with the regression network $f_o$.
More specifically, we use a learning decay on the backbone network to fine-tune the whole network:
\begin{equation}\label{fine-tuning}
    \begin{gathered}
        \theta_b^{'} =  \theta_{b} - \frac{\alpha}{\eta}\nabla_{\theta_{b}}\mathcal{L}, \theta_o^{'} = \theta_{o} - \alpha\nabla_{\theta_{o}}\mathcal{L} \\
        \text{where}~\theta_b = \{\theta^a, \theta^c, \theta^e, \theta^g, \theta^n, \theta^s\}
    \end{gathered}
\end{equation}
where $\mathcal{L}$ is the loss value, $\theta_b$ and $\theta_o$ are the parameters of the backbone network and regression network, respectively, $\alpha$ is the learning rate, $\eta$ is the decay coefficient of learning rate $\alpha$.
\subsection{Optimization Algorithm}
In \model, we consider modeling four categories of features, including adjacent graphs, dynamical GPS reports, inherent POI attributes, and crowd portraits, for POI flow inference.
We construct a spatio-temporal graph encoder to fuse these heterogeneous features and exploit contrastive self-learning to optimize the backbone network with unlabeled data.
At last, we initialize the backbone network with optimal parameters and then fine-tune the network with a smaller learning rate on labeled data.
The detailed process of \model is shown in Algorithm \ref{Alg:model}.
\begin{algorithm}
\fontsize{8}{8}\selectfont
	\LinesNumbered
\KwIn{Adjacent matrix: $\mathcal{A}$; GPS reported users $\mathbf{X}$;\\
\quad\quad\quad inherent attributes $\mathbf{v}^{a}$; crowd portrait $\mathbf{v}^{c}$
}
\KwOut{Inferred crowd flow $\mathbf{Y}$}
{
    \DontPrintSemicolon
	{given unlabeled dataset $\mathcal{D}_{u}$ and labeled dataset $\mathcal{D}_{l}$}\;
	{\tcp*[h]{\model Pre-training on $\mathcal{D}_{u}$}}\;
    generate data augmentations based on key attributes in $\mathbf{v}^{a}$\;
	\Repeat{stopping criteria is met}
	{
	    \tcp*[h]{backbone network forward} \;
		calculate $\mathbf{o}_{t}$ with Eq.(\ref{node_encoder}), (\ref{message_network}), and (\ref{fusion_network})\;
		\tcp*[h] {contrastive representation learning} \;
		forward the prototype network\;
		calculate contrastive loss $\mathcal{L}_{c}$ with Eq.(\ref{main_swap_loss})\;
		back-propagation to optimize $\theta_{b}$\;
	}
    get the learned backbone parameters $\theta_{b}^{*}$\;
    {\tcp*[h]{\model fine-tuning on $\mathcal{D}_{l}$}}\;
    initialize the backbone network with $\theta_{b}^{*}$\;
	\Repeat{stopping criteria is met}
	{
		randomly select a batch from $\mathcal{D}_{l}$\;
		forward the backbone network and $f_o$ on $\mathcal{L}$\;
        backward two networks on $\mathcal{L}$ with Eq.(\ref{fine-tuning})\;
	}
	output inferred flow $\mathbf{Y}$
}
\caption{Algorithm of \model{}}\label{Alg:model}
\end{algorithm} 

\section{Experiments}
In this section, we conduct experiments on two categories of POI flow datasets, including office buildings and residential buildings, to demonstrate the effectiveness of \model. This experiment can be divided into three parts: 1) basic performance experiment, in which the proposed model is compared with existing methods; 2) ablation studies, including the validation of the contrastive self-supervised method and the spatio-temporal representation network; 3) parameter sensitivity analysis. 

\subsection{Experimental Setting}
\subsubsection{Data Description}
We conduct experiments on two real-world POI flow datasets, they are:
\begin{itemize}[leftmargin=*]
    \item \textbf{Office.} This dataset contains the $53,550$ unlabeled offices in China and $1,068$ of them are labeled for supervised learning.
    \item \textbf{Residential.} This dataset contains the GPS reports in residential buildings, which has $237,180$ unlabeled data and $966$ labeled data. 
\end{itemize}

\begin{table}[ht]
	\centering
	\caption{Datasets.}
	\label{tab:dataset}
	\begin{tabular}{ccc}
		\hline
		\textbf{Dataset}& \textbf{Office}
		& \textbf{Residential}    \\
		\hline
        Time & 1/7/2020-31/7/2020 & 1/12/2019-31/12/2019 \\
        Time interval & 1 week & 1 week \\
        \# unlabeled nodes & 53,550 & 237,180\\
        \# labeled nodes & 1,068 & 966\\
        \hline
        \multirow{2}{*} {Attributes} & {area and} & {area, house price and}\\
        & {transportation} & {transportation} \\
        \hline
        \multirow{2}{*} {Portraits} & {age and sex} & {age and sex}\\
        &distributions & distributions \\
		\hline
	\end{tabular}
\end{table}

\begin{figure}[tb]
    \centering
    \includegraphics[width=0.45\linewidth]{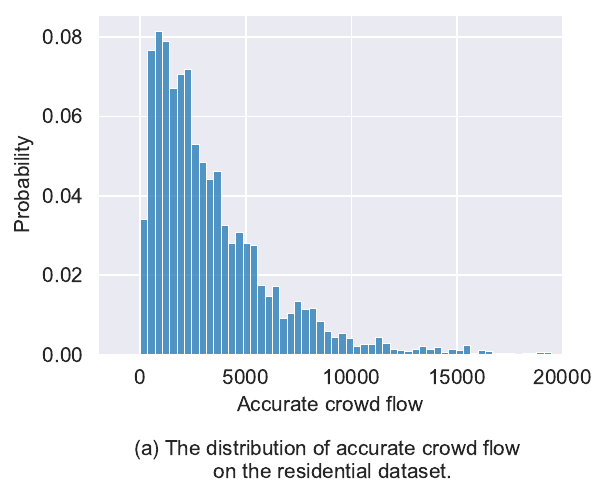}
    \includegraphics[width=0.45\linewidth]{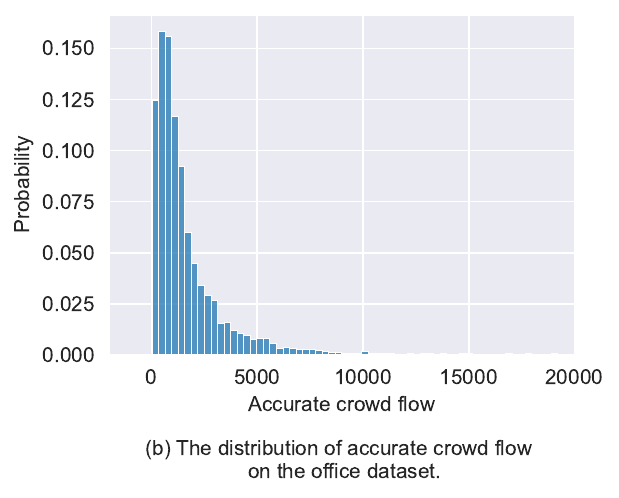}
    \caption{The data distribution on two datasets.}
    \label{fig:dist}
\end{figure}

Figure \ref{fig:dist} show the different distributions of the accurate crowd flow on two datasets. And Table \ref{tab:dataset} shows the statistics of these two datasets. The label's missing ratio in the office dataset is about $2\%$ while that in residential buildings is about $0.4\%$, indicating the missing ratio of labels is quite large.
Moreover, we construct the adjacency graph based on the geospatial distance between POIs and filter out neighbors more than \SI{500}{\meter} away.
For the office dataset, we consider the office attributes in our models, including transportation conditions, age, and gender distribution. And for residential buildings, we consider the additional housing price information, which is an important feature for similar residentials.
Moreover, we focus on inferring the unlabeled POIs from labeled data, not the traditional problem that predicts the future flow from historical observations, which means our task will focus more on capturing the dependence between POIs than between timestamps.
Therefore, we use coarse-grained time data to evaluate our model.

\subsubsection{Evaluation Metrics}
We measure the performance of \model and all baselines by MAPE (Mean Absolute Percentage Error) and ACC (Accuracy):
\begin{equation}\label{metric}
    \text{MAPE}= \frac{1}{n}\sum_{t=1}^{n}|\frac{y_i-\hat{y}_i}{y_i}|
\end{equation}
where $N$ is the number of instances, $\hat{y}_i$ is the inferred flow, and $y_i$ is the ground truth.
Furthermore, for POI $l_i$, we consider it reliable if the inference error (MAPE) of which is less than $0.3$. Otherwise, it is unreliable.
We define the ACC metric as the proportion of reliable instances in Equation.\ref{ACC}.
\begin{equation}\label{ACC}
\text{ACC}= \frac{1}{N}\sum_{i=1}^{N}f(x_i), \ \
  f(x_i)=\begin{cases}
    1, & \text{if $x_i<\epsilon$}.\\
    0, & \text{otherwise}.
  \end{cases}
\end{equation}
where $x_i$ is the MAPE metric, and $\epsilon$ is a relatively small value which is set to $0.3$ in this paper.

\subsubsection{Baseline Algorithms}
We compare the proposed \model with two traditional models and two common deep models for the POI flow inference to verify the performance improvement. The baseline methods are used under the supervised learning setting.
\begin{itemize}[leftmargin=*]
  \item \textbf{LR:} Linear Regression. We train a linear regression model with all kinds of features.
  \item \textbf{XGBoost\cite{xgboost}:} XGBoost is a powerful and widely used ensemble learning model in data mining. We train XGB with all features, which is the same as LR.
  \item \textbf{MLP.} Multi-layer perceptron. We flatten all features and feed them together into the MLP network.
  \item \textbf{MSFNet.} Multi-source fusion network. We divide all features into three categories and feed each category of features into an MLP network to learn the representation.
  Then, we concatenate the embedding and pass them through a fusion network. In other words, MSFNet does not consider the neighbor information and replaces the MPNN component in Figure \ref{method_backbone} with MLP structure.
\end{itemize}


Further, we perform ablation studies to verify the effectiveness of each part, the results are shown in section \ref{ablation_study}.
\begin{itemize}[leftmargin=*]
  \item \textbf{The contrastive self-supervised learning method.} We perform experiments on whether the contrastive self-supervised learning framework is used. To be Specific, comparing the proposed \textbf{\model-Net} with the \textbf{Net} with the contrastive self-supervised learning module, and the \textbf{Net} contains \textbf{MLP}, \textbf{MSFNet}, and the proposed \textbf{STGNN}.
  \item \textbf{The spatio-temporal representation network.} We perform experiments with different backnone networks, including \textbf{MLP}, \textbf{MSFNet}, and the proposed \textbf{STGNN}.
\end{itemize}

\subsubsection{Hyperparameters}
We split the dataset into the train, valid, and test sets, where the proportion of samples in the training set ranges in $[10\%, 20\%, 50\%, 70\%]$. Moreover, we set the validation ratio to $0.1$ and evaluate the model's performance with 10\% - 70\% proportions of samples. Furthermore, we perform 10-fold cross-validation for all models.

In our experiments, for MLP and MSFNet, we first run a grid search on the number of MLP layers $L_m$ ranging in $[1,2,3,4]$ and the hidden dimensions $d$ ranging in $[32, 64, 128, 256, 512, 1024]$ on backbone network, then choose the best hyperparameter settings. For the proposed \model model, in the pre-training stage, there are three kinds of parameters: 1) data augmentation including the number of positive samples $m$; 2) prototype parameters including the number of prototypes $K$, hidden dimensions $d_c$, temperature $\tau$; 3) trainer parameters including learning rate, weight decay, batch size, and max iteration steps. We fix $\tau$ to $0.05$, weight decay to $1e{-4}$ and batch size to $256$. In the fine-tuning stage, we set $\eta$ to $10$ and other parameters the same as the supervised training. For STGNN, we set the hops to $1$ and max neighbors to $20$ for smaller memory usage.

In the ablation studies, for all backbone networks, we search the learning rate $\alpha$ in $[5e^{-3}, 2e^{-3}, 1e^{-3}, 5e^{-4}, 2e^{-4}, 1e^{-4}]$ to find the best and fixed the batch size to $64$. In the parameter sensitivity analysis, we run a grid search on $m$ ranging in $[5, 10, 20]$ and on $d_c$ in $[64, 256, 512]$ to find out the best parameter.
Moreover, all our experiments are conducted on CentOS 7 with a single Tesla V100-PCIE-16GB GPU.

\subsection{Overall performances}
\begin{table*}[!t]
    \centering
    \caption{Performance comparison of different methods on residential and office buildings}\label{Experiments:residential}
    \begin{adjustbox}{width=0.9\linewidth}
    \begin{tabular}{c|cc|cc|cc|cc}
        \Xhline{1.5pt}
        \multirow{2}{*}{\textbf{Residential}}
        &\multicolumn{2}{c|}{\textbf{70\%}}
        & \multicolumn{2}{c|}{\textbf{50\%}}
        &\multicolumn{2}{c|}{\textbf{20\%}}
        & \multicolumn{2}{c}{\textbf{10\%}}
        \\
        \cline{2-9}
         & \textbf{MAPE} & \textbf{ACC}
         & \textbf{MAPE} & \textbf{ACC}
         & \textbf{MAPE} & \textbf{ACC}
         & \textbf{MAPE} & \textbf{ACC}
        \\
        \Xhline{1.2pt}
        LR
        &0.719&0.508
        &0.704&0.493
        &0.690&0.485
        &0.744&0.480
        \\
        XGBoost
        &0.463&0.540
        &0.472&0.542
        &0.497&0.531
        &0.527&0.517
        \\
        \hline
        MLP
        &0.493&0.496
        &0.473&0.465
        &0.539&0.403
        &0.555&0.374
        \\
        MSFNet
        &\textbf{0.373}&\textbf{0.653}
        &\textbf{0.382}&\textbf{0.635}
        &0.401&0.587
        &0.465&0.538
        \\
        \hline
        \emph{CSST}
        &0.389&0.633
        &0.421&0.616
        &\textbf{0.401}&\textbf{0.591}
        &\textbf{0.427}&\textbf{0.580}
        \\
        \Xhline{1.5pt}
        \multirow{2}{*}{\textbf{Office}}
        &\multicolumn{2}{c|}{\textbf{70\%}}
        & \multicolumn{2}{c|}{\textbf{50\%}}
        &\multicolumn{2}{c|}{\textbf{20\%}}
        & \multicolumn{2}{c}{\textbf{10\%}}
        \\
        \cline{2-9}
         & \textbf{MAPE} & \textbf{ACC}
         & \textbf{MAPE} & \textbf{ACC}
         & \textbf{MAPE} & \textbf{ACC}
         & \textbf{MAPE} & \textbf{ACC}
        \\
        \Xhline{1.2pt}
        LR
        &1.035&0.372
        &0.967&0.390
        &0.945&0.404
        &1.024&0.359
        \\
        XGBoost
        &0.470&0.527
        &0.475&0.548
        &0.485&0.535
        &0.507&0.500
        \\
        \hline
        MLP
        &0.451&0.501
        &0.459&0.497
        &0.565&0.474
        &0.521&0.416
        \\
        MSFNet
        &\textbf{0.322}&\textbf{0.620}
        &\textbf{0.329}&\textbf{0.611}
        &0.349&0.590
        &0.365&0.566
        \\\hline
        CSST
        &0.363&0.574
        &0.350&0.605
        &\textbf{0.343}&\textbf{0.606}
        &\textbf{0.356}&\textbf{0.598}
        \\\Xhline{1.5pt}
    \end{tabular}
    \end{adjustbox}
\end{table*}

\begin{figure*}[htbp]
    \centering
    \includegraphics[width=\linewidth]{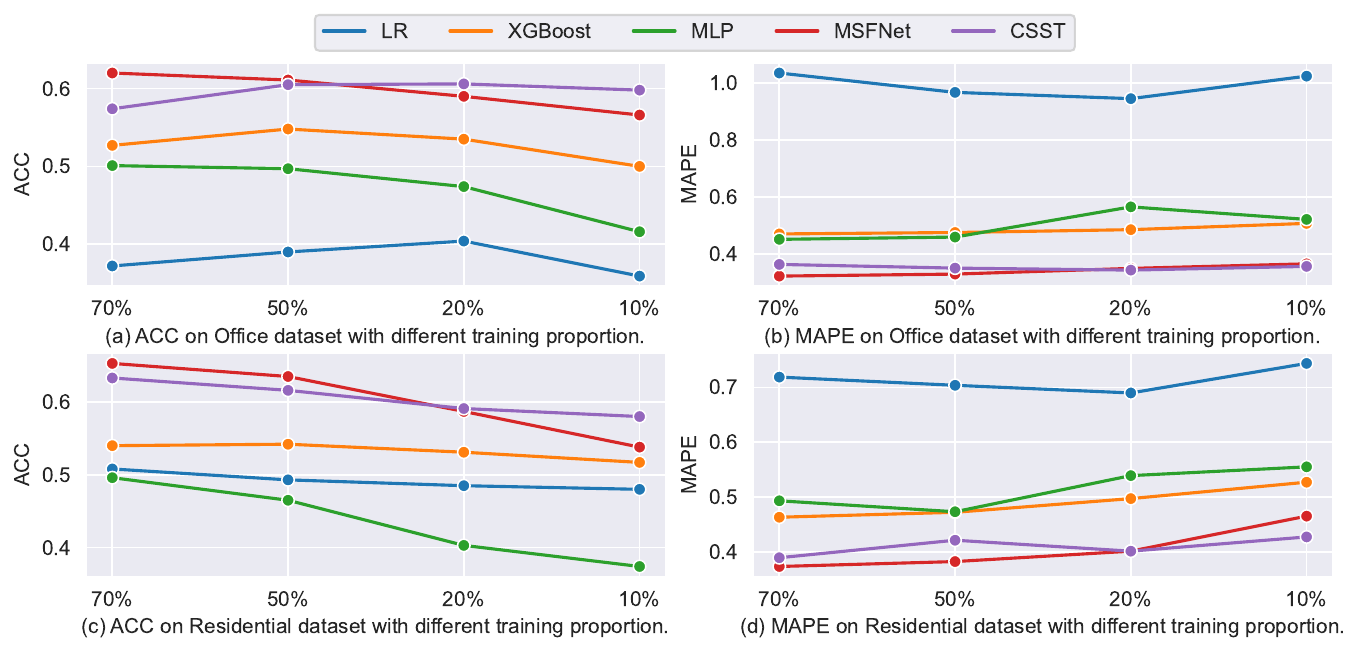}
    \caption{MAPE and ACC on the two datasets with different training proportion.}
    \label{fig:overall}
\end{figure*}

Table \ref{Experiments:residential} and Figure \ref{fig:overall} shows the performance comparison between the proposed model and the baseline methods urder the \textbf{MAPE} and \textbf{ACC} metrics on office and residential buildings.
We have the following observations: 
\begin{enumerate}[leftmargin=*]
    \item For traditional models and simple designed MLP models, they are entirely failed on both datasets. The failure is that features, including low-quality GPS reports, POI attributes, and user portraits, are heterogeneous in structure, and their relationship is complicated.
    \item Compared with a supervised model, our proposed \model framework can improve the performance when the training set is relatively small. Specifically, \model outperforms MSFNet by $4.2\%$ and $3.2\%$ on two datasets in terms of ACC when training proportions are $10\%$, and by $3.8\%$ and $0.9\%$ on two datasets in terms of MAPE.
\end{enumerate}

Besides, with the training data increase, our proposed \model framework performs slightly worse than the baseline supervised model. A possible reason is that the amount of test data is limited when the training proportion increases to $70\%$, which has a similar feature space as the training set, so the pre-trained model empowered by unlabeled data does not contribute to the performance. In our opinion, our proposed \model framework may be more advantageous when the training data is less than testing data.

\subsection{Ablation Studies}\label{ablation_study}
\begin{table*}[!t]
    \centering
    \caption{Ablation experiments results on residential and office buildings}\label{ablation-contrastive:residential}
    \begin{adjustbox}{width=0.9\linewidth}
    \begin{tabular}{c|cc|cc|cc|cc}
        \Xhline{1.5pt}
        \multirow{2}{*}{\textbf{Residential}}
        &\multicolumn{2}{c|}{\textbf{70\%}}
        & \multicolumn{2}{c|}{\textbf{50\%}}
        &\multicolumn{2}{c|}{\textbf{20\%}}
        & \multicolumn{2}{c}{\textbf{10\%}}
        \\
        \cline{2-9}
         & \textbf{MAPE} & \textbf{ACC}
         & \textbf{MAPE} & \textbf{ACC}
         & \textbf{MAPE} & \textbf{ACC}
         & \textbf{MAPE} & \textbf{ACC}
        \\
        \Xhline{1pt}
        MLP
        &0.493&0.496
        &0.473&0.465
        &0.539&0.403
        &0.555&0.374
        \\
        CSST-MLP
        &\textbf{0.481}&\textbf{0.480}
        &\textbf{0.457}&\textbf{0.512}
        &\textbf{0.452}&\textbf{0.492}
        &\textbf{0.445}&\textbf{0.505}
        \\
        \hline
        MSFNet
        &0.465&0.538
        &0.401&0.587
        &0.382&0.635
        &0.373&0.653
        \\
        CSST-MSFNet
        &\textbf{0.367}&0.652
        &\textbf{0.379}&\textbf{0.650}
        &\textbf{0.403}&\textbf{0.615}
        &\textbf{0.410}&\textbf{0.611}
        \\
        \hline
        STGNN
        &\textbf{0.373}&\textbf{0.649}
        &\textbf{0.391}&\textbf{0.628}
        &0.424&\textbf{0.599}
        &0.445&0.579
        \\
        CSST-STGNN
        &0.389&0.633
        &0.421&0.616
        &\textbf{0.401}&0.591
        &\textbf{0.427}&\textbf{0.580}
        \\
        \Xhline{1.5pt}
        \multirow{2}{*}{\textbf{Office}}
        &\multicolumn{2}{c|}{\textbf{70\%}}
        & \multicolumn{2}{c|}{\textbf{50\%}}
        &\multicolumn{2}{c|}{\textbf{20\%}}
        & \multicolumn{2}{c}{\textbf{10\%}}
        \\
        \cline{2-9}
         & \textbf{MAPE} & \textbf{ACC}
         & \textbf{MAPE} & \textbf{ACC}
         & \textbf{MAPE} & \textbf{ACC}
         & \textbf{MAPE} & \textbf{ACC}
        \\
        \Xhline{1pt}
        MLP
        &0.451&0.501
        &0.459&0.497
        &0.565&0.474
        &0.521&0.416
        \\
        CSST-MLP
        &\textbf{0.447}&\textbf{0.527}
        &\textbf{0.441}&\textbf{0.499}
        &\textbf{0.474}&\textbf{0.492}        
        &\textbf{0.478}&\textbf{0.468}
        \\
        \hline
        MSFNet
        &\textbf{0.322}&\textbf{0.620}
        &\textbf{0.329}&\textbf{0.611}
        &0.349&0.590
        &0.365&0.566
        \\
        CSST-MSFNet
        &0.356&0.596
        &0.331&0.606
        &\textbf{0.340}&\textbf{0.607}
        &\textbf{0.353}&\textbf{0.596}
        \\
        \hline
        STGNN
        &0.371&\textbf{0.580}
        &0.380&0.593
        &0.264&0.595
        &0.463&0.578
        \\
        CSST-STGNN
        &\textbf{0.363}&0.574
        &\textbf{0.350}&\textbf{0.605}
        &\textbf{0.343}&\textbf{0.606}
        &\textbf{0.356}&\textbf{0.598}
        \\
        \Xhline{1.5pt}
    \end{tabular}
    \end{adjustbox}
\end{table*}
This section mainly validates the contrastive self-supervised framework and the spatio-temporal representation network (STGNN) of the proposed method. Table \ref{ablation-contrastive:residential} summarize the performance comparison between each backbone network and its enhanced method \model-Net. The observations are as follows: 
\begin{enumerate}[leftmargin=*]
    \item The contrastive self-supervised module significantly improves the effect. To be specific, \model-MLP significantly outperforms MLP by $10.6\%$, $10.9\%$ , $2.7\%$ and $0.9\%$ on residential data in terms of ACC when training proportions are $10\%$, $20\%$, $50\%$ and $70\%$ respectively. \model-MSFNet significantly outperforms MSFNet by $7.3\%$, $2.8\%$ and $1.5\%$ on residential data in terms of ACC when training proportions are $10\%$, $20\%$ and $50\%$. \model-STGNN has empowered STGNN by $2.0\%$, $1.1\%$ on office data in terms of ACC with $10\%$ and $20\%$ training samples.
    \item The proposed STGNN outperforms the baseline backbone networks. Specifically, STGNN significantly outperforms MSFNet by $4.1\%$ and $1.2\%$ on residential data in terms of ACC when training proportions are $10\%$ and $20\%$. It can be seen that, as the training proportion increases, STGNN performs slightly worse than MSFNet. We think the graph-based method will be more affected by the labeled topology imbalance.
\end{enumerate}

\begin{figure*}[t]
    \centering
    \includegraphics[width=\linewidth]{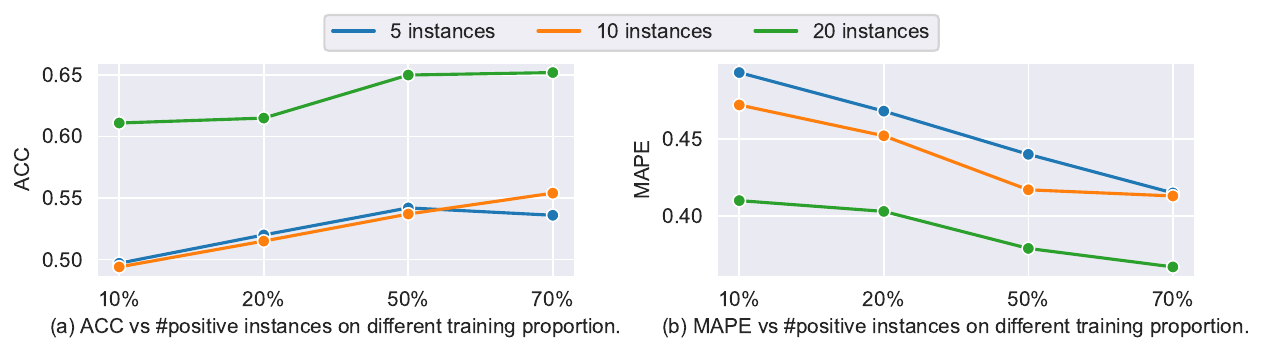}
    \includegraphics[width=\linewidth]{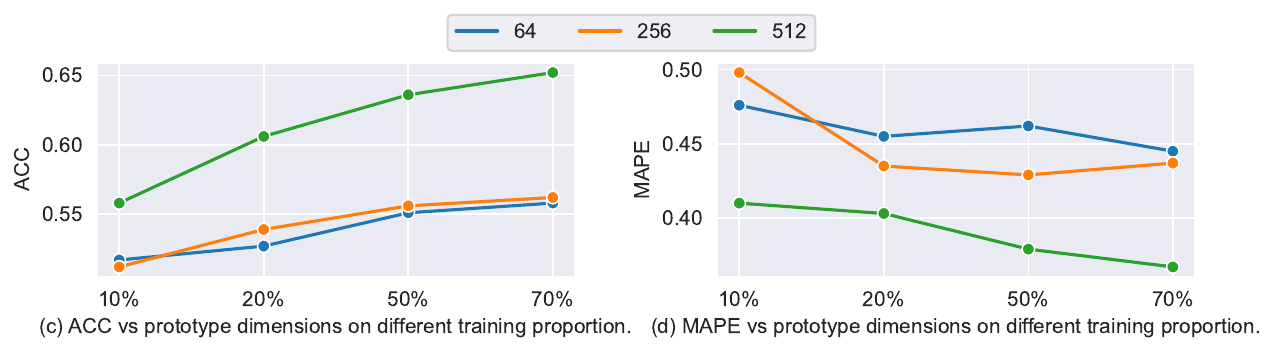}
    \caption{Studies on hyperparameters. (a)(b) show the model performance under different settings of the number of positive instances; (c)(d) show the model performance under different numbers of prototype dimensions.}\label{exper:hyper_param}
\end{figure*}

\subsection{Parameter Sensitivity Analysis}\label{parameter_sensitivity}
There are many important hyper-parameters for the proposed \model at both the pre-training and fine-tuning stages.
In this paper, we focus on the parameters of the pre-training stage, including the number of positive samples $m$ and the hidden dimension of the prototype network $d_c$,
while keeping the parameter settings at the fine-tuning stage the same as supervised learning models.
Specifically, to investigate the robustness of \model, we compare the \model empowered MSFNet on $m$ and $d_c$ on the residential dataset.

We first evaluate the impact of $m$ with the $d_c$ fixed to $512$ by default and $m$ vary in $[5, 10, 20]$, the results are shown in Figure \ref{exper:hyper_param}.
We can observe that the inference accuracy increases obviously as $m$ increases in terms of both metrics, and the model achieves the best performance when $m=20$.
In addition, the results show that although the performance setting of $m=10$ slightly performs better than $m=5$, both models perform even worse than the supervised model,
which demonstrates that the success of \model{} requires large-scale positive data augmentations.
However, the computation time increases linearly as the number of samples increases, so we set the $m$ to a maximum of $20$ to ensure the model converges as fast as possible.

Furthermore, to evaluate the impact of $d_c$, we fixed the number of positive samples $m$ to $20$ by default and vary $d_c$ in $[64, 256, 512]$, the results are shown in Figure \ref{exper:hyper_param}(c) and \ref{exper:hyper_param}(d).
We can observe that \model-MSFNet failed on settings $64$ and $256$ while achieving the best results when $d_c=512$,
which indicates the effectiveness of contrastive learning framework benefits to large prototype dimensions.
Even more, \model-MSFNet performs worse than supervised learning models, which demonstrates that the hidden dimension of the prototype network plays a decisive role in the success of our model.

\section{Related Work}
\subsection{Deep Learning on Spatio-Temporal Data}
Spatio-temporal data has been widely used in smart cities. The works related to this paper can be roughly divided into three categories, spatio-temporal forecasting, spatio-temporal data imputation, and spatio-temporal inference. 

Spatio-temporal \emph{forecasting} aims to use historical data to \textbf{predict future flows}. Recently, with the rapid development of deep learning, many researchers have devoted themselves to forecasting spatio-temporal data with deep models \cite{STResNet,STGNN2018,ijcai2020flow,STGCN01,STGCN02}. 

Spatio-temporal data \emph{imputation} focuses on \textbf{imputing missing values} in the historical data. Traditionally, missing values are randomly generated at the data acquisition of sensors due to device or transmission failure. Many researchers have studied data imputation problems under different settings \cite{asadi2019convolution,gao2020generative,yoon2018estimating}.

Spatio-temporal \emph{inference} hopes to \textbf{infer the true value} through incomplete observations, which means we can only collect a part of the true distribution. \cite{Cheng2018A,Han2021Fine} proposes deep models to infer the air quality at unobserved locations. \cite{urbanfmkdd20} tries to formulate the fine-grained grid flow inference as a high-resolution image recovery problem and address it with a convolution network associated with the distributional up-sampling module. 

Unlike the existing methods, this paper aims to infer the unobserved POI flows with large-scale noisy data and a small amount of accurate flow data. Specifically, most POIs do not have accurate historical observations, only noise flow information, and few POIs have accurate flow data. Therefore, we try to design a suitable transfer learning method to improve the model's generalization ability and reduce the impact of overfitting.

\subsection{Self-Supervised Learning}
As we all know, training effective deep networks usually rely on large-scale labeled data, but collecting accurate data is expensive.
To address this problem, many self-supervised learning methods \cite{liu2020selfsupervised,CSL2018google,jaiswal2021survey} have been proposed to learn the representation from large-scale unlabeled data.
Self-supervised learning models can be divided into two categories: contrastive and generative approaches.
The generative approaches \cite{GAN01,GAN02,GAN03,autoencoder01,autoencoder02,autoencoder03} define the reconstruction loss to recovery the input instance with probability distributions.
The contrastive learning method \cite{chen2020csl,misra2019selfsupervised,xie2021self} calculates the pairwise similarity in a representation space, and its performance depends on the data augmentation methods and the number of negative samples.
The main idea of contrastive learning methods is that similar samples extracted from different augmented data methods should be close to each other in the representation space, and dissimilar samples should be far away.
Authors proposed to exploit large and consistent dictionaries learned with contrastive loss to improve the performance \cite{he2019moco,chen2020improved}.
In addition, the work \cite{chen2020csl} shows that the memory bank can be entirely replaced with a large batch where other elements in the same batch can be negative instances.
However, both methods are memory inefficient because they require a larger batch size and additional momentum encoder.
To avoid large memory occupation, researchers proposed a swapped prediction \cite{caron2021swaped}. They predict the code of one view from the representation of another view, which is simpler than previous methods and achieves state-of-the-art performance in image classification tasks.
Furthermore, the great success of contrastive self-learning has inspired researchers \cite{qiu2020gcc,you2020graph,zhu2020deep,zhu2021graph,liu2021spatio} to explore the application to graph neural networks.

Moreover, there are also some works about self-supervisd learning on spatio-temporal data. UrbanSTC\cite{qu2022forecasting} utilizes spatial and temporal self-supervision to better forecasting fine-grained crowd flow data from coarse-grained spatio-temporal data. ST-SSL\cite{ji2023spatio} involves clustering algorithms to produce better self-supervising signals and achieves better performance on the grid-based traffic flow prediction problem. However, inferring POI-level crowd flow from low-quality data is still challenging. Therefore,  we first try to design the ST data augmentation based on a swapped contrastive learning framework tailored for POI-level crowd flow inference tasks.

\section{Conclusion}
In this paper, we introduce a novel contrastive self-learning framework, designated as \model, specifically developed for the purpose of fine-grained spatio-temporal flow inference. \model encompasses a data augmentation component, which partitions the attributes into several subsets, and samples similar instances within the same set. It further adopts a swapped-based contrastive learning methodology to foster the learning of transferable representations for POIs. Our experiments substantiate the effectiveness of the graph neural network in modeling intricate spatial dependencies. Furthermore, we assess the performance of \model in conjunction with three diverse backbone networks, ranging from MLP to graph neural networks, and illustrate that \model can be integrated with varying deep learning models to consistently enhance performance.

In future research, we plan to delve into addressing the issue of low-quality spatio-temporal data for prediction tasks through the application of contrastive learning.

\bibliographystyle{ACM-Reference-Format}
\bibliography{ref}


\begin{thebibliography}{47}


\ifx \showCODEN    \undefined \def \showCODEN     #1{\unskip}     \fi
\ifx \showDOI      \undefined \def \showDOI       #1{#1}\fi
\ifx \showISBNx    \undefined \def \showISBNx     #1{\unskip}     \fi
\ifx \showISBNxiii \undefined \def \showISBNxiii  #1{\unskip}     \fi
\ifx \showISSN     \undefined \def \showISSN      #1{\unskip}     \fi
\ifx \showLCCN     \undefined \def \showLCCN      #1{\unskip}     \fi
\ifx \shownote     \undefined \def \shownote      #1{#1}          \fi
\ifx \showarticletitle \undefined \def \showarticletitle #1{#1}   \fi
\ifx \showURL      \undefined \def \showURL       {\relax}        \fi
\providecommand\bibfield[2]{#2}
\providecommand\bibinfo[2]{#2}
\providecommand\natexlab[1]{#1}
\providecommand\showeprint[2][]{arXiv:#2}

\bibitem[Asadi and Regan(2019)]%
        {asadi2019convolution}
\bibfield{author}{\bibinfo{person}{Reza Asadi} {and} \bibinfo{person}{Amelia
  Regan}.} \bibinfo{year}{2019}\natexlab{}.
\newblock \showarticletitle{A convolution recurrent autoencoder for
  spatio-temporal missing data imputation}.
\newblock \bibinfo{journal}{\emph{arXiv preprint arXiv:1904.12413}}
  (\bibinfo{year}{2019}).
\newblock


\bibitem[Caron et~al\mbox{.}(2021)]%
        {caron2021swaped}
\bibfield{author}{\bibinfo{person}{Mathilde Caron}, \bibinfo{person}{Ishan
  Misra}, \bibinfo{person}{Julien Mairal}, \bibinfo{person}{Priya Goyal},
  \bibinfo{person}{Piotr Bojanowski}, {and} \bibinfo{person}{Armand Joulin}.}
  \bibinfo{year}{2021}\natexlab{}.
\newblock \bibinfo{title}{Unsupervised Learning of Visual Features by
  Contrasting Cluster Assignments}.
\newblock
\newblock
\showeprint[arxiv]{2006.09882}~[cs.CV]


\bibitem[Chen et~al\mbox{.}(2019)]%
        {STGCN02}
\bibfield{author}{\bibinfo{person}{Cen Chen}, \bibinfo{person}{Kenli Li},
  \bibinfo{person}{Sin~G. Teo}, \bibinfo{person}{Xiaofeng Zou},
  \bibinfo{person}{kang Wang}, \bibinfo{person}{jie Wang}, {and}
  \bibinfo{person}{Zeng Zeng}.} \bibinfo{year}{2019}\natexlab{}.
\newblock \showarticletitle{Gated Residual Recurrent Graph Neural Networks for
  Traffic Prediction}. In \bibinfo{booktitle}{\emph{Proceedings of the
  Thirty-Third AAAI Conference on Artificial Intelligence (AAAI-19)}}.
\newblock


\bibitem[Chen and Guestrin(2016)]%
        {xgboost}
\bibfield{author}{\bibinfo{person}{Tianqi Chen} {and} \bibinfo{person}{Carlos
  Guestrin}.} \bibinfo{year}{2016}\natexlab{}.
\newblock \showarticletitle{XGBoost}.
\newblock \bibinfo{journal}{\emph{Proceedings of the 22nd ACM SIGKDD
  International Conference on Knowledge Discovery and Data Mining}}
  (\bibinfo{date}{Aug} \bibinfo{year}{2016}).
\newblock
\showISBNx{9781450342322}
\urldef\tempurl%
\url{https://doi.org/10.1145/2939672.2939785}
\showDOI{\tempurl}


\bibitem[Chen et~al\mbox{.}(2020b)]%
        {chen2020csl}
\bibfield{author}{\bibinfo{person}{Ting Chen}, \bibinfo{person}{Simon
  Kornblith}, \bibinfo{person}{Mohammad Norouzi}, {and}
  \bibinfo{person}{Geoffrey Hinton}.} \bibinfo{year}{2020}\natexlab{b}.
\newblock \showarticletitle{A simple framework for contrastive learning of
  visual representations}. In \bibinfo{booktitle}{\emph{International
  conference on machine learning}}. PMLR, \bibinfo{pages}{1597--1607}.
\newblock


\bibitem[Chen et~al\mbox{.}(2020c)]%
        {chen2020bigcsl}
\bibfield{author}{\bibinfo{person}{Ting Chen}, \bibinfo{person}{Simon
  Kornblith}, \bibinfo{person}{Kevin Swersky}, \bibinfo{person}{Mohammad
  Norouzi}, {and} \bibinfo{person}{Geoffrey Hinton}.}
  \bibinfo{year}{2020}\natexlab{c}.
\newblock \showarticletitle{Big Self-Supervised Models are Strong
  Semi-Supervised Learners}.
\newblock \bibinfo{journal}{\emph{arXiv preprint arXiv:2006.10029}}
  (\bibinfo{year}{2020}).
\newblock


\bibitem[Chen et~al\mbox{.}(2020a)]%
        {chen2020improved}
\bibfield{author}{\bibinfo{person}{Xinlei Chen}, \bibinfo{person}{Haoqi Fan},
  \bibinfo{person}{Ross Girshick}, {and} \bibinfo{person}{Kaiming He}.}
  \bibinfo{year}{2020}\natexlab{a}.
\newblock \showarticletitle{Improved baselines with momentum contrastive
  learning}.
\newblock \bibinfo{journal}{\emph{arXiv preprint arXiv:2003.04297}}
  (\bibinfo{year}{2020}).
\newblock


\bibitem[Cheng et~al\mbox{.}(2018)]%
        {Cheng2018A}
\bibfield{author}{\bibinfo{person}{Weiyu Cheng}, \bibinfo{person}{Yanyan Shen},
  \bibinfo{person}{Yanmin Zhu}, {and} \bibinfo{person}{Linpeng Huang}.}
  \bibinfo{year}{2018}\natexlab{}.
\newblock \showarticletitle{{A Neural Attention Model for Urban Air Quality
  Inference: Learning the Weights of Monitoring Stations}}.
\newblock \bibinfo{journal}{\emph{Proceedings of the AAAI Conference on
  Artificial Intelligence}} \bibinfo{volume}{32}, \bibinfo{number}{1}
  (\bibinfo{date}{apr} \bibinfo{year}{2018}).
\newblock
\showISSN{2374-3468}
\urldef\tempurl%
\url{https://ojs.aaai.org/index.php/AAAI/article/view/11871}
\showURL{%
\tempurl}


\bibitem[Devlin et~al\mbox{.}(2019)]%
        {devlin2018bert}
\bibfield{author}{\bibinfo{person}{Jacob Devlin}, \bibinfo{person}{Ming-Wei
  Chang}, \bibinfo{person}{Kenton Lee}, {and} \bibinfo{person}{Kristina
  Toutanova}.} \bibinfo{year}{2019}\natexlab{}.
\newblock \showarticletitle{BERT: Pre-training of Deep Bidirectional
  Transformers for Language Understanding}. In
  \bibinfo{booktitle}{\emph{NAACL-HLT (1)}}.
\newblock


\bibitem[Donahue et~al\mbox{.}(2017)]%
        {GAN03}
\bibfield{author}{\bibinfo{person}{Jeff Donahue}, \bibinfo{person}{Philipp
  Krähenbühl}, {and} \bibinfo{person}{Trevor Darrell}.}
  \bibinfo{year}{2017}\natexlab{}.
\newblock \bibinfo{title}{Adversarial Feature Learning}.
\newblock
\newblock
\showeprint[arxiv]{1605.09782}~[cs.LG]


\bibitem[Donahue and Simonyan(2019)]%
        {GAN02}
\bibfield{author}{\bibinfo{person}{Jeff Donahue} {and} \bibinfo{person}{Karen
  Simonyan}.} \bibinfo{year}{2019}\natexlab{}.
\newblock \showarticletitle{Large Scale Adversarial Representation Learning}.
\newblock \bibinfo{journal}{\emph{Advances in Neural Information Processing
  Systems}}  \bibinfo{volume}{32} (\bibinfo{year}{2019}),
  \bibinfo{pages}{10542--10552}.
\newblock


\bibitem[Feng et~al\mbox{.}(2020)]%
        {ijcai2020flow}
\bibfield{author}{\bibinfo{person}{Jie Feng}, \bibinfo{person}{Ziqian Lin},
  \bibinfo{person}{Tong Xia}, \bibinfo{person}{Funing Sun},
  \bibinfo{person}{Diansheng Guo}, {and} \bibinfo{person}{Yong Li}.}
  \bibinfo{year}{2020}\natexlab{}.
\newblock \showarticletitle{A Sequential Convolution Network for Population
  Flow Prediction with Explicitly Correlation Modelling}. In
  \bibinfo{booktitle}{\emph{Proceedings of the Twenty-Ninth International Joint
  Conference on Artificial Intelligence, {IJCAI-20}}},
  \bibfield{editor}{\bibinfo{person}{Christian Bessiere}} (Ed.).
  \bibinfo{publisher}{International Joint Conferences on Artificial
  Intelligence Organization}, \bibinfo{pages}{1331--1337}.
\newblock
\urldef\tempurl%
\url{https://doi.org/10.24963/ijcai.2020/185}
\showDOI{\tempurl}
\newblock
\shownote{Main track}.


\bibitem[Gao et~al\mbox{.}(2020)]%
        {gao2020generative}
\bibfield{author}{\bibinfo{person}{Nan Gao}, \bibinfo{person}{Hao Xue},
  \bibinfo{person}{Wei Shao}, \bibinfo{person}{Sichen Zhao},
  \bibinfo{person}{Kyle~Kai Qin}, \bibinfo{person}{Arian Prabowo},
  \bibinfo{person}{Mohammad~Saiedur Rahaman}, {and} \bibinfo{person}{Flora~D
  Salim}.} \bibinfo{year}{2020}\natexlab{}.
\newblock \showarticletitle{Generative adversarial networks for spatio-temporal
  data: A survey}.
\newblock \bibinfo{journal}{\emph{arXiv preprint arXiv:2008.08903}}
  (\bibinfo{year}{2020}).
\newblock


\bibitem[Gilmer et~al\mbox{.}(2017)]%
        {mpnn2017icml}
\bibfield{author}{\bibinfo{person}{Justin Gilmer}, \bibinfo{person}{Samuel~S
  Schoenholz}, \bibinfo{person}{Patrick~F Riley}, \bibinfo{person}{Oriol
  Vinyals}, {and} \bibinfo{person}{George~E Dahl}.}
  \bibinfo{year}{2017}\natexlab{}.
\newblock \showarticletitle{Neural message passing for quantum chemistry}. In
  \bibinfo{booktitle}{\emph{International conference on machine learning}}.
  PMLR, \bibinfo{pages}{1263--1272}.
\newblock


\bibitem[Goodfellow et~al\mbox{.}(2014)]%
        {GAN01}
\bibfield{author}{\bibinfo{person}{Ian Goodfellow}, \bibinfo{person}{Jean
  Pouget-Abadie}, \bibinfo{person}{Mehdi Mirza}, \bibinfo{person}{Bing Xu},
  \bibinfo{person}{David Warde-Farley}, \bibinfo{person}{Sherjil Ozair},
  \bibinfo{person}{Aaron Courville}, {and} \bibinfo{person}{Yoshua Bengio}.}
  \bibinfo{year}{2014}\natexlab{}.
\newblock \showarticletitle{Generative adversarial nets}.
\newblock \bibinfo{journal}{\emph{Advances in neural information processing
  systems}}  \bibinfo{volume}{27} (\bibinfo{year}{2014}).
\newblock


\bibitem[Han et~al\mbox{.}(2021)]%
        {Han2021Fine}
\bibfield{author}{\bibinfo{person}{Qilong Han}, \bibinfo{person}{Dan Lu}, {and}
  \bibinfo{person}{Rui Chen}.} \bibinfo{year}{2021}\natexlab{}.
\newblock \showarticletitle{{Fine-Grained Air Quality Inference via
  Multi-Channel Attention Model}}.
\newblock  (\bibinfo{year}{2021}), \bibinfo{pages}{2512--2518}.
\newblock
\urldef\tempurl%
\url{https://doi.org/10.24963/ijcai.2021/346}
\showDOI{\tempurl}


\bibitem[He et~al\mbox{.}(2020)]%
        {he2019moco}
\bibfield{author}{\bibinfo{person}{Kaiming He}, \bibinfo{person}{Haoqi Fan},
  \bibinfo{person}{Yuxin Wu}, \bibinfo{person}{Saining Xie}, {and}
  \bibinfo{person}{Ross Girshick}.} \bibinfo{year}{2020}\natexlab{}.
\newblock \showarticletitle{Momentum contrast for unsupervised visual
  representation learning}. In \bibinfo{booktitle}{\emph{Proceedings of the
  IEEE/CVF Conference on Computer Vision and Pattern Recognition}}.
  \bibinfo{pages}{9729--9738}.
\newblock


\bibitem[Jaiswal et~al\mbox{.}(2021)]%
        {jaiswal2021survey}
\bibfield{author}{\bibinfo{person}{Ashish Jaiswal},
  \bibinfo{person}{Ashwin~Ramesh Babu}, \bibinfo{person}{Mohammad~Zaki Zadeh},
  \bibinfo{person}{Debapriya Banerjee}, {and} \bibinfo{person}{Fillia
  Makedon}.} \bibinfo{year}{2021}\natexlab{}.
\newblock \showarticletitle{A survey on contrastive self-supervised learning}.
\newblock \bibinfo{journal}{\emph{Technologies}} \bibinfo{volume}{9},
  \bibinfo{number}{1} (\bibinfo{year}{2021}), \bibinfo{pages}{2}.
\newblock


\bibitem[Ji et~al\mbox{.}(2023)]%
        {ji2023spatio}
\bibfield{author}{\bibinfo{person}{Jiahao Ji}, \bibinfo{person}{Jingyuan Wang},
  \bibinfo{person}{Chao Huang}, \bibinfo{person}{Junjie Wu},
  \bibinfo{person}{Boren Xu}, \bibinfo{person}{Zhenhe Wu},
  \bibinfo{person}{Junbo Zhang}, {and} \bibinfo{person}{Yu Zheng}.}
  \bibinfo{year}{2023}\natexlab{}.
\newblock \showarticletitle{Spatio-temporal self-supervised learning for
  traffic flow prediction}. In \bibinfo{booktitle}{\emph{Proceedings of the
  AAAI Conference on Artificial Intelligence}}, Vol.~\bibinfo{volume}{37}.
  \bibinfo{pages}{4356--4364}.
\newblock


\bibitem[Liang et~al\mbox{.}(2019)]%
        {urbanfmkdd20}
\bibfield{author}{\bibinfo{person}{Yuxuan Liang}, \bibinfo{person}{Kun Ouyang},
  \bibinfo{person}{Lin Jing}, \bibinfo{person}{Sijie Ruan}, \bibinfo{person}{Ye
  Liu}, \bibinfo{person}{Junbo Zhang}, \bibinfo{person}{David~S. Rosenblum},
  {and} \bibinfo{person}{Yu Zheng}.} \bibinfo{year}{2019}\natexlab{}.
\newblock \showarticletitle{UrbanFM: Inferring Fine-Grained Urban Flows}.
\newblock \bibinfo{journal}{\emph{CoRR}}  \bibinfo{volume}{abs/1902.05377}
  (\bibinfo{year}{2019}).
\newblock
\showeprint[arxiv]{1902.05377}
\urldef\tempurl%
\url{http://arxiv.org/abs/1902.05377}
\showURL{%
\tempurl}


\bibitem[Lin et~al\mbox{.}(2019)]%
        {lin2019deepstn}
\bibfield{author}{\bibinfo{person}{Ziqian Lin}, \bibinfo{person}{Jie Feng},
  \bibinfo{person}{Ziyang Lu}, \bibinfo{person}{Yong Li}, {and}
  \bibinfo{person}{Depeng Jin}.} \bibinfo{year}{2019}\natexlab{}.
\newblock \showarticletitle{DeepSTN+: Context-Aware Spatial-Temporal Neural
  Network for Crowd Flow Prediction in Metropolis}.
\newblock \bibinfo{journal}{\emph{Proceedings of the AAAI Conference on
  Artificial Intelligence}} \bibinfo{volume}{33}, \bibinfo{number}{01}
  (\bibinfo{date}{Jul.} \bibinfo{year}{2019}), \bibinfo{pages}{1020--1027}.
\newblock
\urldef\tempurl%
\url{https://doi.org/10.1609/aaai.v33i01.33011020}
\showDOI{\tempurl}


\bibitem[Liu et~al\mbox{.}(2021a)]%
        {liu2021spatio}
\bibfield{author}{\bibinfo{person}{Xu Liu}, \bibinfo{person}{Yuxuan Liang},
  \bibinfo{person}{Yu Zheng}, \bibinfo{person}{Bryan Hooi}, {and}
  \bibinfo{person}{Roger Zimmermann}.} \bibinfo{year}{2021}\natexlab{a}.
\newblock \showarticletitle{Spatio-Temporal Graph Contrastive Learning}.
\newblock \bibinfo{journal}{\emph{arXiv preprint arXiv:2108.11873}}
  (\bibinfo{year}{2021}).
\newblock


\bibitem[Liu et~al\mbox{.}(2021b)]%
        {liu2020selfsupervised}
\bibfield{author}{\bibinfo{person}{Xiao Liu}, \bibinfo{person}{Fanjin Zhang},
  \bibinfo{person}{Zhenyu Hou}, \bibinfo{person}{Li Mian},
  \bibinfo{person}{Zhaoyu Wang}, \bibinfo{person}{Jing Zhang}, {and}
  \bibinfo{person}{Jie Tang}.} \bibinfo{year}{2021}\natexlab{b}.
\newblock \showarticletitle{Self-supervised learning: Generative or
  contrastive}.
\newblock \bibinfo{journal}{\emph{IEEE Transactions on Knowledge and Data
  Engineering}} (\bibinfo{year}{2021}).
\newblock


\bibitem[Misra and Maaten(2020)]%
        {misra2019selfsupervised}
\bibfield{author}{\bibinfo{person}{Ishan Misra} {and} \bibinfo{person}{Laurens
  van~der Maaten}.} \bibinfo{year}{2020}\natexlab{}.
\newblock \showarticletitle{Self-supervised learning of pretext-invariant
  representations}. In \bibinfo{booktitle}{\emph{Proceedings of the IEEE/CVF
  Conference on Computer Vision and Pattern Recognition}}.
  \bibinfo{pages}{6707--6717}.
\newblock


\bibitem[Pathak et~al\mbox{.}(2016)]%
        {autoencoder02}
\bibfield{author}{\bibinfo{person}{Deepak Pathak}, \bibinfo{person}{Philipp
  Kr{\"a}henb{\"u}hl}, \bibinfo{person}{Jeff Donahue}, \bibinfo{person}{Trevor
  Darrell}, {and} \bibinfo{person}{Alexei~A Efros}.}
  \bibinfo{year}{2016}\natexlab{}.
\newblock \showarticletitle{Context Encoders: Feature Learning by Inpainting}.
  In \bibinfo{booktitle}{\emph{2016 IEEE Conference on Computer Vision and
  Pattern Recognition (CVPR)}}. IEEE, \bibinfo{pages}{2536--2544}.
\newblock


\bibitem[Qiu et~al\mbox{.}(2020)]%
        {qiu2020gcc}
\bibfield{author}{\bibinfo{person}{Jiezhong Qiu}, \bibinfo{person}{Qibin Chen},
  \bibinfo{person}{Yuxiao Dong}, \bibinfo{person}{Jing Zhang},
  \bibinfo{person}{Hongxia Yang}, \bibinfo{person}{Ming Ding},
  \bibinfo{person}{Kuansan Wang}, {and} \bibinfo{person}{Jie Tang}.}
  \bibinfo{year}{2020}\natexlab{}.
\newblock \bibinfo{booktitle}{\emph{GCC: Graph Contrastive Coding for Graph
  Neural Network Pre-Training}}.
\newblock \bibinfo{publisher}{Association for Computing Machinery},
  \bibinfo{address}{New York, NY, USA}, \bibinfo{pages}{1150–1160}.
\newblock
\showISBNx{9781450379984}
\urldef\tempurl%
\url{https://doi.org/10.1145/3394486.3403168}
\showURL{%
\tempurl}


\bibitem[Qu et~al\mbox{.}(2022)]%
        {qu2022forecasting}
\bibfield{author}{\bibinfo{person}{Hao Qu}, \bibinfo{person}{Yongshun Gong},
  \bibinfo{person}{Meng Chen}, \bibinfo{person}{Junbo Zhang},
  \bibinfo{person}{Yu Zheng}, {and} \bibinfo{person}{Yilong Yin}.}
  \bibinfo{year}{2022}\natexlab{}.
\newblock \showarticletitle{Forecasting fine-grained urban flows via
  spatio-temporal contrastive self-supervision}.
\newblock \bibinfo{journal}{\emph{IEEE Transactions on Knowledge and Data
  Engineering}} (\bibinfo{year}{2022}).
\newblock


\bibitem[Radford(2018)]%
        {GPT01}
\bibfield{author}{\bibinfo{person}{A. Radford}.}
  \bibinfo{year}{2018}\natexlab{}.
\newblock \showarticletitle{Improving Language Understanding by Generative
  Pre-Training}.
\newblock


\bibitem[Radford et~al\mbox{.}(2019)]%
        {LM01}
\bibfield{author}{\bibinfo{person}{A. Radford}, \bibinfo{person}{Jeffrey Wu},
  \bibinfo{person}{R. Child}, \bibinfo{person}{David Luan},
  \bibinfo{person}{Dario Amodei}, {and} \bibinfo{person}{Ilya Sutskever}.}
  \bibinfo{year}{2019}\natexlab{}.
\newblock \showarticletitle{Language Models are Unsupervised Multitask
  Learners}.
\newblock


\bibitem[Sun et~al\mbox{.}(2020)]%
        {junkai2020}
\bibfield{author}{\bibinfo{person}{Junkai Sun}, \bibinfo{person}{Junbo Zhang},
  \bibinfo{person}{Qiaofei Li}, \bibinfo{person}{Xiuwen Yi},
  \bibinfo{person}{Yuxuan Liang}, {and} \bibinfo{person}{Yu Zheng}.}
  \bibinfo{year}{2020}\natexlab{}.
\newblock \showarticletitle{Predicting Citywide Crowd Flows in Irregular
  Regions Using Multi-View Graph Convolutional Networks}.
\newblock \bibinfo{journal}{\emph{IEEE Transactions on Knowledge and Data
  Engineering}}  \bibinfo{volume}{PP} (\bibinfo{date}{07}
  \bibinfo{year}{2020}), \bibinfo{pages}{1--1}.
\newblock
\urldef\tempurl%
\url{https://doi.org/10.1109/TKDE.2020.3008774}
\showDOI{\tempurl}


\bibitem[Tian et~al\mbox{.}(2020)]%
        {mview2019contrastive}
\bibfield{author}{\bibinfo{person}{Yonglong Tian}, \bibinfo{person}{Dilip
  Krishnan}, {and} \bibinfo{person}{Phillip Isola}.}
  \bibinfo{year}{2020}\natexlab{}.
\newblock \showarticletitle{Contrastive multiview coding}. In
  \bibinfo{booktitle}{\emph{Computer Vision--ECCV 2020: 16th European
  Conference, Glasgow, UK, August 23--28, 2020, Proceedings, Part XI 16}}.
  Springer, \bibinfo{pages}{776--794}.
\newblock


\bibitem[van~den Oord et~al\mbox{.}(2018)]%
        {CSL2018google}
\bibfield{author}{\bibinfo{person}{Aaron van~den Oord}, \bibinfo{person}{Yazhe
  Li}, {and} \bibinfo{person}{Oriol Vinyals}.} \bibinfo{year}{2018}\natexlab{}.
\newblock \showarticletitle{Representation Learning with Contrastive Predictive
  Coding}.
\newblock \bibinfo{journal}{\emph{arXiv preprint arXiv:1807.03748v2}}
  (\bibinfo{year}{2018}).
\newblock


\bibitem[Vincent et~al\mbox{.}(2008)]%
        {autoencoder01}
\bibfield{author}{\bibinfo{person}{Pascal Vincent}, \bibinfo{person}{Hugo
  Larochelle}, \bibinfo{person}{Y. Bengio}, {and}
  \bibinfo{person}{Pierre-Antoine Manzagol}.} \bibinfo{year}{2008}\natexlab{}.
\newblock \showarticletitle{Extracting and composing robust features with
  denoising autoencoders}.
\newblock \bibinfo{journal}{\emph{Proceedings of the 25th International
  Conference on Machine Learning}}, \bibinfo{pages}{1096--1103}.
\newblock
\urldef\tempurl%
\url{https://doi.org/10.1145/1390156.1390294}
\showDOI{\tempurl}


\bibitem[Wang et~al\mbox{.}(2018)]%
        {glue2018}
\bibfield{author}{\bibinfo{person}{Alex Wang}, \bibinfo{person}{Amanpreet
  Singh}, \bibinfo{person}{Julian Michael}, \bibinfo{person}{Felix Hill},
  \bibinfo{person}{Omer Levy}, {and} \bibinfo{person}{Samuel~R. Bowman}.}
  \bibinfo{year}{2018}\natexlab{}.
\newblock \showarticletitle{{GLUE:} {A} Multi-Task Benchmark and Analysis
  Platform for Natural Language Understanding}.
\newblock \bibinfo{journal}{\emph{CoRR}}  \bibinfo{volume}{abs/1804.07461}
  (\bibinfo{year}{2018}).
\newblock
\showeprint[arxiv]{1804.07461}
\urldef\tempurl%
\url{http://arxiv.org/abs/1804.07461}
\showURL{%
\tempurl}


\bibitem[Wu et~al\mbox{.}(2018)]%
        {wu2018unsupervised}
\bibfield{author}{\bibinfo{person}{Zhirong Wu}, \bibinfo{person}{Yuanjun
  Xiong}, \bibinfo{person}{X~Yu Stella}, {and} \bibinfo{person}{Dahua Lin}.}
  \bibinfo{year}{2018}\natexlab{}.
\newblock \showarticletitle{Unsupervised Feature Learning via Non-Parametric
  Instance Discrimination}. In \bibinfo{booktitle}{\emph{Proceedings of the
  IEEE Conference on Computer Vision and Pattern Recognition}}.
\newblock


\bibitem[Xie et~al\mbox{.}(2021)]%
        {xie2021self}
\bibfield{author}{\bibinfo{person}{Zhenda Xie}, \bibinfo{person}{Yutong Lin},
  \bibinfo{person}{Zhuliang Yao}, \bibinfo{person}{Zheng Zhang},
  \bibinfo{person}{Qi Dai}, \bibinfo{person}{Yue Cao}, {and}
  \bibinfo{person}{Han Hu}.} \bibinfo{year}{2021}\natexlab{}.
\newblock \showarticletitle{Self-supervised learning with swin transformers}.
\newblock \bibinfo{journal}{\emph{arXiv preprint arXiv:2105.04553}}
  (\bibinfo{year}{2021}).
\newblock


\bibitem[Xu et~al\mbox{.}(2016)]%
        {li2016mflow}
\bibfield{author}{\bibinfo{person}{Fengli Xu}, \bibinfo{person}{Pengyu~Zhang
  profile}, {and} \bibinfo{person}{Yong Li}.} \bibinfo{year}{2016}\natexlab{}.
\newblock \showarticletitle{Context-aware real-time population estimation for
  metropolis}. In \bibinfo{booktitle}{\emph{Proceedings of the 2016 ACM
  International Joint Conference on Pervasive and Ubiquitous Computing
  (Ubicomp-16)}}. \bibinfo{pages}{1064--1075}.
\newblock


\bibitem[Yoon et~al\mbox{.}(2018)]%
        {yoon2018estimating}
\bibfield{author}{\bibinfo{person}{Jinsung Yoon}, \bibinfo{person}{William~R
  Zame}, {and} \bibinfo{person}{Mihaela van~der Schaar}.}
  \bibinfo{year}{2018}\natexlab{}.
\newblock \showarticletitle{Estimating missing data in temporal data streams
  using multi-directional recurrent neural networks}.
\newblock \bibinfo{journal}{\emph{IEEE Transactions on Biomedical Engineering}}
  \bibinfo{volume}{66}, \bibinfo{number}{5} (\bibinfo{year}{2018}),
  \bibinfo{pages}{1477--1490}.
\newblock


\bibitem[You et~al\mbox{.}(2020)]%
        {you2020graph}
\bibfield{author}{\bibinfo{person}{Yuning You}, \bibinfo{person}{Tianlong
  Chen}, \bibinfo{person}{Yongduo Sui}, \bibinfo{person}{Ting Chen},
  \bibinfo{person}{Zhangyang Wang}, {and} \bibinfo{person}{Yang Shen}.}
  \bibinfo{year}{2020}\natexlab{}.
\newblock \showarticletitle{Graph contrastive learning with augmentations}.
\newblock \bibinfo{journal}{\emph{Advances in Neural Information Processing
  Systems}}  \bibinfo{volume}{33} (\bibinfo{year}{2020}),
  \bibinfo{pages}{5812--5823}.
\newblock


\bibitem[Yu et~al\mbox{.}(2017a)]%
        {STGCN01}
\bibfield{author}{\bibinfo{person}{Bing Yu}, \bibinfo{person}{Haoteng Yin},
  {and} \bibinfo{person}{Zhanxing Zhu}.} \bibinfo{year}{2017}\natexlab{a}.
\newblock \showarticletitle{Spatio-Temporal Graph Convolutional Networks: A
  Deep Learning Framework for Traffic Forecasting}. In
  \bibinfo{booktitle}{\emph{Proceedings of the Twenty-sixth International Joint
  Conference on Artificial Intelligence, {IJCAI-17}}}.
\newblock


\bibitem[Yu et~al\mbox{.}(2017b)]%
        {STGNN2018}
\bibfield{author}{\bibinfo{person}{Bing Yu}, \bibinfo{person}{Haoteng Yin},
  {and} \bibinfo{person}{Zhanxing Zhu}.} \bibinfo{year}{2017}\natexlab{b}.
\newblock \showarticletitle{Spatio-temporal Graph Convolutional Neural Network:
  {A} Deep Learning Framework for Traffic Forecasting}.
\newblock \bibinfo{journal}{\emph{CoRR}}  \bibinfo{volume}{abs/1709.04875}
  (\bibinfo{year}{2017}).
\newblock
\showeprint[arxiv]{1709.04875}
\urldef\tempurl%
\url{http://arxiv.org/abs/1709.04875}
\showURL{%
\tempurl}


\bibitem[Zhang et~al\mbox{.}(2017b)]%
        {STResNet}
\bibfield{author}{\bibinfo{person}{Junbo Zhang}, \bibinfo{person}{Yu Zheng},
  {and} \bibinfo{person}{Dekang Qi}.} \bibinfo{year}{2017}\natexlab{b}.
\newblock \showarticletitle{Deep Spatio-Temporal Residual Networks for Citywide
  Crowd Flows Prediction}. In \bibinfo{booktitle}{\emph{Proceedings of the
  Thirty-First AAAI Conference on Artificial Intelligence (AAAI-17)}}.
  \bibinfo{pages}{1655--1661}.
\newblock


\bibitem[Zhang et~al\mbox{.}(2017a)]%
        {autoencoder03}
\bibfield{author}{\bibinfo{person}{Richard Zhang}, \bibinfo{person}{Phillip
  Isola}, {and} \bibinfo{person}{Alexei~A Efros}.}
  \bibinfo{year}{2017}\natexlab{a}.
\newblock \showarticletitle{Split-brain autoencoders: Unsupervised learning by
  cross-channel prediction}. In \bibinfo{booktitle}{\emph{Proceedings of the
  IEEE Conference on Computer Vision and Pattern Recognition}}.
  \bibinfo{pages}{1058--1067}.
\newblock


\bibitem[Zheng(2019)]%
        {zheng2019urban}
\bibfield{author}{\bibinfo{person}{Yu Zheng}.} \bibinfo{year}{2019}\natexlab{}.
\newblock \bibinfo{booktitle}{\emph{Urban computing}}.
\newblock \bibinfo{publisher}{MIT Press}.
\newblock


\bibitem[Zheng et~al\mbox{.}(2014)]%
        {zheng2014urban}
\bibfield{author}{\bibinfo{person}{Yu Zheng}, \bibinfo{person}{Licia Capra},
  \bibinfo{person}{Ouri Wolfson}, {and} \bibinfo{person}{Hai Yang}.}
  \bibinfo{year}{2014}\natexlab{}.
\newblock \showarticletitle{Urban Computing: Concepts, Methodologies, and
  Applications}.
\newblock \bibinfo{journal}{\emph{ACM Transaction on Intelligent Systems and
  Technology}} (\bibinfo{date}{October} \bibinfo{year}{2014}).
\newblock


\bibitem[Zhu et~al\mbox{.}(2020)]%
        {zhu2020deep}
\bibfield{author}{\bibinfo{person}{Yanqiao Zhu}, \bibinfo{person}{Yichen Xu},
  \bibinfo{person}{Feng Yu}, \bibinfo{person}{Qiang Liu}, \bibinfo{person}{Shu
  Wu}, {and} \bibinfo{person}{Liang Wang}.} \bibinfo{year}{2020}\natexlab{}.
\newblock \showarticletitle{Deep graph contrastive representation learning}.
\newblock \bibinfo{journal}{\emph{arXiv preprint arXiv:2006.04131}}
  (\bibinfo{year}{2020}).
\newblock


\bibitem[Zhu et~al\mbox{.}(2021)]%
        {zhu2021graph}
\bibfield{author}{\bibinfo{person}{Yanqiao Zhu}, \bibinfo{person}{Yichen Xu},
  \bibinfo{person}{Feng Yu}, \bibinfo{person}{Qiang Liu}, \bibinfo{person}{Shu
  Wu}, {and} \bibinfo{person}{Liang Wang}.} \bibinfo{year}{2021}\natexlab{}.
\newblock \showarticletitle{Graph contrastive learning with adaptive
  augmentation}. In \bibinfo{booktitle}{\emph{Proceedings of the Web Conference
  2021}}. \bibinfo{pages}{2069--2080}.
\newblock


\end{thebibliography}

\end{document}